\documentclass{article} 
\usepackage{iclr2027_conference,times}
\iclrfinalcopy

\usepackage{amsmath,amsfonts,bm}

\def\eqref#1{equation~\ref{#1}}

\def\1{\bm{1}}

\DeclareMathAlphabet{\mathsfit}{\encodingdefault}{\sfdefault}{m}{sl}
\SetMathAlphabet{\mathsfit}{bold}{\encodingdefault}{\sfdefault}{bx}{n}

\usepackage{url}
\usepackage{amsmath}
\usepackage{algorithm}
\usepackage{algpseudocode}
\usepackage{booktabs}
\usepackage{multirow}
\usepackage{graphicx}
\usepackage{enumitem}
\usepackage{wrapfig}

\usepackage{xcolor} 
\usepackage[table]{xcolor}
\definecolor{links}{HTML}{0078b0} 
\definecolor{files}{HTML}{fc6160}

\usepackage[
  colorlinks=true,            
  linkcolor=files,
  filecolor=files,
  urlcolor=links,
  citecolor=links
]{hyperref}

\title{CoMoFlow: Controllable Protein Conformational Ensemble Generation via Explicit Collective-Mode Modeling}

\author{%
\textbf{Yusong Wang$^{1}$, Zheyuan Gu$^{2}$,
Ailiang Lin$^{1}$, Minghao Shao$^{3}$} \\
\textbf{Yicheng Xu$^{1}$, Jiang Cai$^{4}$,
Prayag Tiwari$^{5}$, Mingkun Xu$^{4}$} \\[0.5em]
$^{1}$ Beijing University of Chinese Medicine, \quad
$^{2}$ Peking University, \quad
$^{3}$ New York University \\
$^{4}$ Guangdong Institute of Intelligence Science and Technology \quad
$^{5}$ Halmstad University College
}

\begin{document}

\maketitle
\lhead{Preprint}

\begin{abstract}
Protein conformational ensemble generation aims to efficiently characterize the structural variations accessible to a protein beyond a single static structure.
Recent flow-based methods efficiently generate conformational ensembles, but they model coordinated structural variation implicitly, which can limit ensemble fidelity and hinder direct control.
To address this limitation, we propose \textbf{CoMoFlow}, a controllable framework for generating protein backbone conformational ensembles by explicitly modeling collective variation.
CoMoFlow constructs protein-specific collective modes from the equilibrium structure, learns the distribution of their coefficients across the molecular dynamics (MD) ensemble, and uses sampled collective variations to guide SE(3) flow matching through feature conditioning and prior recentering.
Experiments on ATLAS and de novo proteins show that CoMoFlow improves reconstruction of MD conformational distributions and coordinated structural variation relative to strong flow-based baselines, while enabling direct control over targeted conformational changes.
\end{abstract}

\section{Introduction}
Proteins are inherently dynamic molecules whose functions often depend on structural variations across a conformational ensemble rather than on a single static structure\citep{henzler2007dynamic,boehr2009role}.
Molecular dynamics (MD) simulations characterize such conformational ensembles at atomic resolution by sampling protein conformations along dynamic trajectories.
The resulting MD ensembles reveal accessible protein conformations and provide valuable information for applications including structure-based drug design, virtual screening, and protein engineering \citep{childers2017insights,sledz2018protein,amaro2010emerging}.
However, conventional MD simulations require integration with extremely small time steps over long trajectories, requiring many computationally expensive force evaluations for adequate conformational sampling \citep{hollingsworth2018molecular}.
Recent deep generative models provide a more efficient alternative by learning MD conformational distributions and directly generating diverse conformations without explicitly simulating the full MD trajectory \citep{jing2024alphafold,sengar2026generative}.

\begin{figure}[t]
\includegraphics[width=0.98\textwidth]{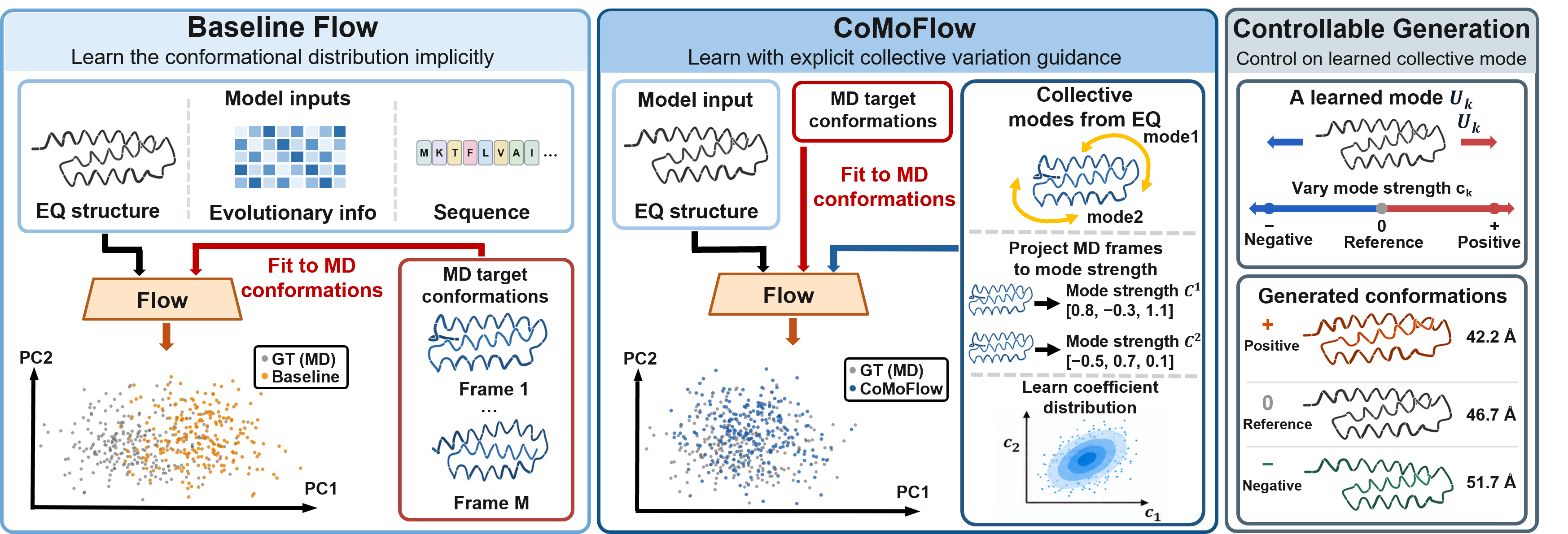}
\centering
\caption{Baseline flow matching and CoMoFlow on the ATLAS protein 7asgA.
Baseline methods learn from individual MD conformations, whereas CoMoFlow learns collective modes from the equilibrium structure and their variation across the MD ensemble to guide generation.
Manipulating selected modes enables controllable conformational changes.
\AA{} denotes inter-region distances.
}
\label{first_fig}\vspace{-0.5cm}
\end{figure}

Flow matching has emerged as an effective framework for protein conformational ensemble generation \citep{lipman2023flow,wang2509simplefold}. 
By learning a continuous vector field that transports samples from a simple prior distribution toward the target conformational distribution, flow-based models provide an efficient alternative to stochastic reverse-diffusion sampling \citep{tong2023improving}.
Early approaches such as AlphaFlow and ESMFlow repurpose pretrained structure predictors for ensemble generation, with their template-conditioned variants further incorporating a reference structure \citep{jing2024alphafold}.
AlphaFlow-Lit \citep{li2024improving} reduces the computational burden of the pretrained-model pipeline, while P2DFlow \citep{jin2025p2dflow} and IDPFold2 \citep{zhu2026extending} introduce enhanced flow formulations and representations for modeling more complex conformational distributions.
A complementary direction is to avoid dependence on large pretrained folding models altogether.
BBFlow \citep{wolf2025conformational} conditions both the flow and its prior on an equilibrium backbone structure without relying on pretrained folding models or evolutionary information.

Existing flow-based methods can efficiently generate diverse conformations by learning from MD ensembles.
These ensembles capture coordinated structural variation across residues \citep{garcia1992large,amadei1993essential}.
Flow matching learns the target distribution through conditional transport paths defined for individual samples \citep{lipman2023flow}. 
In existing protein flow models, individual MD conformations serve as training targets, whereas ensemble-level coordinated variation is represented implicitly in the learned distribution.
When such coordinated variation is not fully recovered, generated ensembles may misrepresent both the relative populations of conformational states and ensemble-level patterns of structural variation, as illustrated in the left part of \autoref{first_fig}.

To address this limitation, we propose \textbf{CoMoFlow}, which augments flow-based conformational ensemble generation with an explicit mechanism for modeling collective variation across the ensemble.
As illustrated in the middle part of \autoref{first_fig}, CoMoFlow constructs protein-specific collective modes from the equilibrium structure and projects MD conformations onto these modes to learn the distribution of their coefficients across the ensemble.
During generation, a sampled coefficient vector defines a collective displacement field that guides an SE(3) flow-matching generator through feature conditioning and translational prior recentering.
This explicit collective representation not only improves the recovery of ensemble-level structural variation but also enables controllable generation.
As illustrated in the right part of \autoref{first_fig}, modifying selected mode coefficients steers generated conformations along the corresponding collective modes, enabling targeted exploration of structural changes with control over both direction and magnitude.
Contributions are summarized as follows:
\begin{itemize}[leftmargin=*]
    \item We propose CoMoFlow, which introduces explicit modeling of ensemble-level coordinated conformational variation into flow-based protein conformational ensemble generation, enabling more faithful reconstruction of the conformational ensemble and its coordinated structural variation.

    \item CoMoFlow further enables controllable conformation generation, allowing generated conformations to be steered along selected collective modes with controllable direction and magnitude, thereby supporting targeted exploration of specific conformational changes.

    \item Experiments show that CoMoFlow better reconstructs MD conformational distributions and coordinated variation than strong flow-based baselines, while enabling reliable controllable generation.
\end{itemize}

\section{Related Work}

\subsection{Flow-based Protein Conformational Ensemble Generation}
Conventional MD \citep{karplus2002molecular} samples conformational dynamics through numerical integration, while Replica-Exchange Molecular Dynamics \citep{sugita1999replica} enhances sampling by exchanging configurations among replicas.
To reduce computational cost, deep generative models directly learn conformational distributions, including latent and temperature-conditioned diffusion models, score-based methods, and flow-matching approaches \citep{zheng2024predicting,lu2024str2str,jing2024alphafold,janson2025deep}.
AlphaFlow and ESMFlow adapt pretrained AlphaFold and ESMFold models, respectively, for conformational ensemble generation, while AlphaFlow-T and ESMFlow-T further incorporate a reference structure \citep{jing2024alphafold}.
AlphaFlow-Lit \citep{li2024improving} reduces the computational cost of AlphaFlow-based pipeline.
P2DFlow \citep{jin2025p2dflow} combines SE(3) flow matching with an informative structural prior and ensemble-level physical information.
IDPFold2 \citep{zhu2026extending} incorporates a mixture-of-experts architecture into flow matching to model heterogeneous conformational behavior.
Complementary to methods built on pretrained folding models, BBFlow \citep{wolf2025conformational} conditions both the flow and its prior on an equilibrium backbone structure without pretrained folding models or evolutionary information.
Despite these different designs, coordinated variation across the ensemble remains largely implicit in the learned generative distribution \citep{bafna2026learning}.

\subsection{Collective Conformational Modeling}
Collective conformational modeling has long been used to characterize large-scale structural variation in proteins \citep{brooks1983harmonic,tama2001conformational,alexandrov2005normal,yang2007well}. 
Normal mode analysis and elastic network models use local harmonic approximations around a reference structure to derive low-frequency collective modes capturing cooperative structural changes \citep{tirion1996large,bahar1997efficient,hinsen1998analysis}.
Trajectory-based approaches such as principal component analysis and essential dynamics instead extract dominant correlated variations from MD ensembles, providing low-dimensional descriptions of conformational changes \citep{amadei1993essential,hayward1995harmonicity,amadei1996efficient,kitao1998energy}.
Learning-based approaches such as SeaMoon and PETIMOT predict protein-specific collective representations from sequence or structure \citep{lombard2025seamoon,lombard2026petimot}.
However, existing collective modeling methods are not directly suited to ensemble generation.
NMA and ENM derive collective modes from local harmonic approximations around a reference structure \citep{levitt1985protein,petrone2006can}, whereas PCA and essential dynamics require pre-existing ensembles \citep{bakan2011prody}.
Learning-based methods predict collective representations from sequence or structure but do not directly generate ensembles.
Motivated by structure-based analyses of collective tendencies and evidence that equilibrium backbone geometry effectively conditions ensemble generation \citep{tirion1996large,atilgan2001anisotropy,wolf2025conformational}, we learn protein-specific collective representations from the equilibrium structure for conformational ensemble generation. 


\section{Methodology}
\begin{figure}[t]
\includegraphics[width=0.95\textwidth]{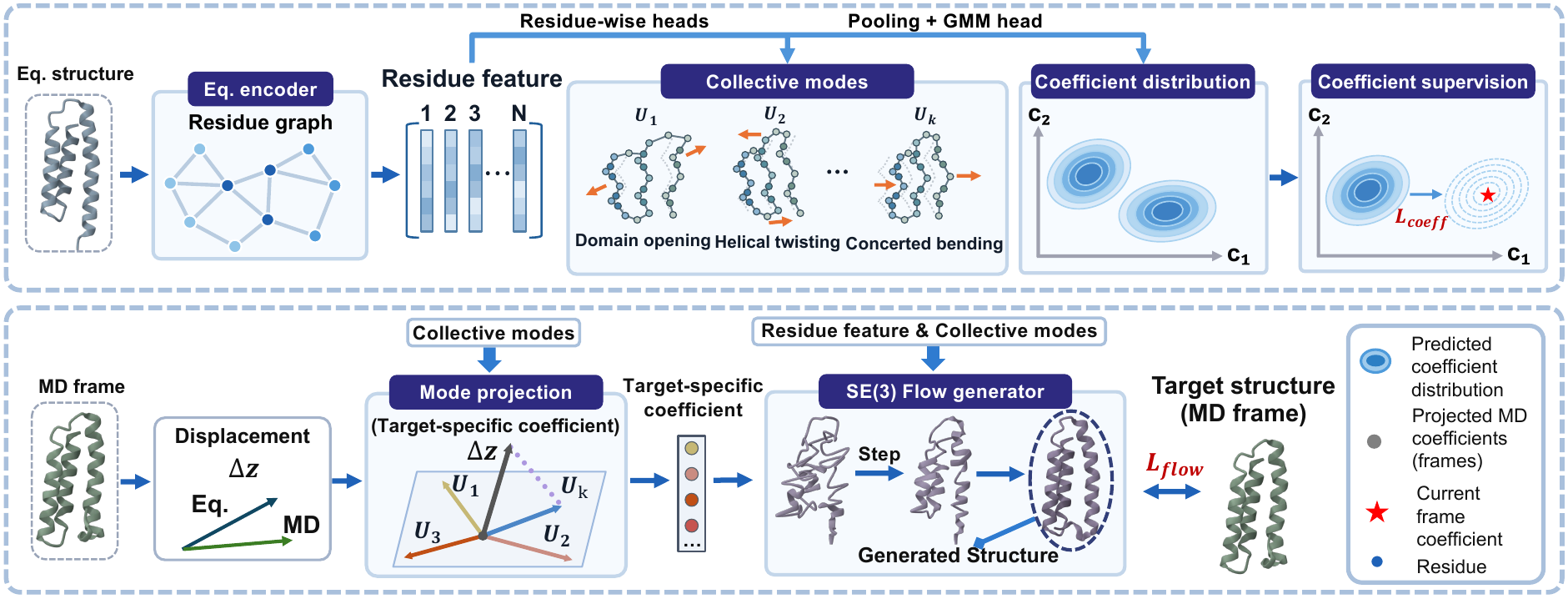}
\centering
\caption{Overview of the CoMoFlow framework, consisting of equilibrium encoding, collective-mode learning and coefficient distribution modeling, and coefficient-guided SE(3) flow generation.}
\label{model_structure}
\vspace{-5mm}
\end{figure}

CoMoFlow generates conformational ensembles from a single equilibrium structure through three main components, as illustrated in \autoref{model_structure}.

\subsection{Problem Definition}
\label{sec:problem}
Given a single equilibrium structure $x_{\mathrm{eq}}$ of a protein as input, our goal is to generate a conformational ensemble that faithfully matches the target MD ensemble.
Let $x_{\mathrm{eq}}=\{(r_i,z_i)\}_{i=1}^{N}\in\mathrm{SE}(3)^N$ denote the equilibrium backbone structure of a protein with $N$ residues, where each residue is represented by a frame consisting of a rotation $r_i\in\mathrm{SO}(3)$ and a translation $z_i\in\mathbb{R}^3$.
CoMoFlow therefore models backbone conformational ensembles at the backbone level.
We denote the target conformational distribution by $p^\ast(x\mid x_{\mathrm{eq}})$ and the available MD ensemble by $\mathcal{D}_{\mathrm{MD}}=\{x_{\mathrm{MD}}^{(j)}\}_{j=1}^{T}$, whose conformations serve as training targets.
Formally, we learn a conditional generative model $\mathcal{M}:x_{\mathrm{eq}}\mapsto\{x^{(m)}\}_{m=1}^{M}$ to approximate the target distribution $p^\ast(x\mid x_{\mathrm{eq}})$.

\subsection{Collective Conformational Modes}
We explicitly represent these coordinated variations using $K$ protein-specific collective conformational modes $U=\{U_k\}_{k=1}^{K}$.
The collective modes explicitly model backbone translations, while residue orientations are handled by the equivariant SE(3) flow under collective conditioning.
Since protein conformational changes often involve coordinated motions across residues \citep{amadei1993essential,bahar1997efficient}, each mode $U_k$ is represented as a residue-wise vector field encoding one such collective pattern \citep{tirion1996large}.
For a specific conformation, the modes are combined linearly using a coefficient vector $c=(c_1,\ldots,c_K)$, where each $c_k$ controls the contribution of mode $U_k$.
The corresponding collective displacement field is:
\begin{equation}
d(c)=\sum_{k=1}^{K}c_kU_k.
\end{equation}
Unlike harmonic normal modes, the learned modes are optimized with MD supervision, and their linear combination serves as collective guidance for the nonlinear SE(3) flow.

We construct each protein-specific mode $U_k$ from the equilibrium structure $x_{\mathrm{eq}}$.
We first encode the equilibrium structure into residue-level representations capturing local structural context.
Specifically, a structure encoder $E_{\theta}$ takes $x_{\mathrm{eq}}$ as input and produces a hidden representation $h_i\in\mathbb{R}^{D}$ for each residue $i$, collectively denoted as $h=\{h_i\}_{i=1}^{N}\in\mathbb{R}^{N\times D}$.
Each mode $U_k$ is parameterized using an equivariant geometric field whose rotational and translational components are shared across residues.
We center the equilibrium coordinates as $\tilde{z}_i=z_i-\bar{z}$, where $\bar{z}=\sum_i w_i z_i/\sum_i w_i$ and $w_i$ denotes the residue-validity mask for residue $i$.
From the rotation-invariant residue representations $h$, the network predicts mode-specific signed scalar weights
$s_{k,i}^{\mathrm{rot}}$ and $s_{k,i}^{\mathrm{trans}}$ over the residues.
We construct a rotational vector and translational component as:
\begin{equation}
\bar{\omega}_k= \frac{\sum_i s_{k,i}^{\mathrm{rot}}\tilde{z}_i} {\sum_i |s_{k,i}^{\mathrm{rot}}|+\epsilon},
\qquad
\omega_k = \rho_k\frac{\bar{\omega}_k} {\|\bar{\omega}_k\|_2+\epsilon},
\qquad
v_k= \frac{\sum_i s_{k,i}^{\mathrm{trans}}\tilde{z}_i} {\sum_i |s_{k,i}^{\mathrm{trans}}|+\epsilon},
\end{equation}
where $\rho_k\geq 0$ is a dimensionless mode-specific scalar predicted from the pooled invariant representation.
The shared residue-wise field for mode $k$ is then:
\begin{equation}
b_{i,k}=\omega_k\times\tilde{z}_i+v_k.
\end{equation}
Here, $\omega_k$ is dimensionless, while $\tilde{z}i$ and $v_k$ have units of length, ensuring dimensional consistency in $b{i,k}$.
We fix the rotational origin at the protein centroid, as any rotation center can be absorbed into the translational component:
$\omega_k\times(\tilde{z}_i-o_k)+v_k
=\omega_k\times\tilde{z}_i+(v_k-\omega_k\times o_k)$.
Because $s_{k,i}^{\mathrm{rot}}$, $s_{k,i}^{\mathrm{trans}}$, and $\rho_k$
are rotation-invariant scalars, under any global rotation $Q\in\mathrm{SO}(3)$, $\omega_k$ and $v_k$ transform as $Q\omega_k$ and $Qv_k$, respectively. 
Consequently, $b_{i,k}\mapsto Qb_{i,k}$, establishing rotation equivariance.

To capture residue-specific deviations while preserving equivariance, we predict the residual in each residue's local frame.
For each residue $i$ and mode $k$, a linear head maps the rotation-invariant representation $h_i$ to local-frame components
$\tilde{\delta}_{i,k}\in\mathbb{R}^3$.
Using the residue orientation $r_i\in\mathrm{SO}(3)$, we transform the residual to the global frame and combine it with the shared geometric field as:
\begin{equation}
\delta_{i,k}=r_i\tilde{\delta}_{i,k},
\qquad
\tilde U_{i,k}=\beta_k b_{i,k}+\alpha\delta_{i,k},
\end{equation}
$\beta_k$ is a learned mode-specific scale, while $\alpha$ is a fixed hyperparameter controlling the residual contribution.
To obtain a nonredundant orthonormal basis, we apply masked orthonormalization:
\begin{equation}
U=\operatorname{Orth}_w(\tilde U),\qquad
\langle U_k,U_l\rangle_w=\delta_{kl}.
\end{equation}
Here, the masked inner product is defined as $\langle A,B\rangle_w=\sum_i w_i\langle A_i,B_i\rangle$, so that orthonormalization is performed over the valid residues of each protein.
The resulting $U=\{U_k\}_{k=1}^{K}$ forms the final set of protein-specific collective conformational modes.

\subsection{Conformational Mode Coefficient Distribution}
The collective modes $U$ define protein-specific directions of coordinated structural variation, but do not specify how those directions are expressed across the conformational ensemble. 
For a particular conformation, a coefficient vector $c$ determines the contribution of each mode. 
We therefore model a protein-specific joint distribution $q(c\mid x_{\mathrm{eq}})$ over the mode coefficients to characterize ensemble-level collective variation. This distribution captures both the variation of individual mode contributions and their statistical dependencies among them across the ensemble.

As the conformational ensemble may exhibit multiple patterns of collective variation, its coefficient vectors may follow a multimodal distribution.
We therefore model $q(c\mid x_{\mathrm{eq}})$ using a $C$-component Gaussian mixture whose parameters are predicted from a pooled representation $\bar{h}$, obtained from the residue-level representations $h=\{h_i\}_{i=1}^{N}$.
Specifically, we define: 
\begin{equation}
q(c\mid x_{\mathrm{eq}}) = \sum_{j=1}^{C}\pi_j(\bar{h})\,
\mathcal{N}\!\left(c;\mu_j(\bar{h}),\Sigma_j(\bar{h})\right),
\end{equation}
where $\pi_j(\bar h) \in [0,1]$, $\mu_j(\bar h) \in \mathbb{R}^K$, and
$\Sigma_j(\bar h) \in \mathbb{R}^{K\times K}$ are the mixture weight,
mean vector, and covariance matrix of component $j$, respectively.
Although the collective modes are geometrically orthonormal, their coefficients may remain statistically dependent. We therefore use a full covariance $\Sigma_j(\bar h)=L_j(\bar h)L_j(\bar h)^\top$ to capture cross-mode dependencies, where $L_j(\bar h)$ is lower triangular.
Mixture components need not correspond one-to-one to discrete conformational states and $C$ controls the capacity of the coefficient distribution.
To learn the coefficient distribution, we project each MD conformation $x$ onto the learned collective modes $U$ to obtain $c^\ast(x)$.
For an aligned MD conformation $x$, let $\Delta z=\{\Delta z_i\}_{i=1}^{N}$, with $\Delta z_i=z_i-z_{\mathrm{eq},i}$,
denote its residue-wise translational displacement from the equilibrium structure:
\begin{equation}
c^\ast(x) = \arg\min_{c}
\left\|
\Delta z-\sum_{k=1}^{K}c_kU_k
\right\|_w^2.
\end{equation}
Because the modes are orthonormal under the masked inner product, $c_k^\ast(x)=\langle \Delta z,U_k\rangle_w$.
To ensure that the learned mode subspace captures MD structural variation, we reconstruct each translational displacement from its projected coefficients and minimize the residual outside the subspace:
\begin{equation}
\mathcal{L}_{\mathrm{rec}} = 
\frac{1}{T}
\sum_{t=1}^{T}
\left\|
\Delta z^{(t)} -
\sum_{k=1}^{K}
c_k^\ast(x_{\mathrm{MD}}^{(t)})U_k
\right\|_w^2.
\end{equation}
This equilibrium-referenced objective makes the learned mode subspace capture both the equilibrium-to-ensemble-mean displacement and centered conformational fluctuations, while $\mathcal{L}_{\mathrm{ens}}$ below specifically matches the covariance of the centered projected coefficients.
We fit $q(c\mid x_{\mathrm{eq}})$ to these projected coefficients using the negative log-likelihood:
\begin{equation}
\mathcal{L}_{\mathrm{coeff}} = -\frac{1}{T}
\sum_{t=1}^{T}
\log q\!\left(c^\ast(x_{\mathrm{MD}}^{(t)})\mid x_{\mathrm{eq}}\right).
\end{equation}
At inference time, we sample a coefficient vector $c\sim q(c\mid x_{\mathrm{eq}})$ and combine it with the collective modes to obtain a conformation-specific displacement field $d(c)=\sum_{k=1}^{K}c_kU_k$.
In addition to coefficient samples, the mixture distribution provides an analytic covariance:
\begin{equation}
\Gamma = \operatorname{Cov}_{q(c\mid x_{\mathrm{eq}})}[c] =
\sum_{j=1}^{C}
\pi_j(\bar h)
\left(
\Sigma_j(\bar h)+\mu_j(\bar h)\mu_j(\bar h)^{\top}
\right)
-\bar{\mu}\bar{\mu}^{\top},
\qquad
\bar{\mu}=\sum_{j=1}^{C}\pi_j(\bar h)\mu_j(\bar h),
\end{equation}
which summarizes the joint variation of the mode coefficients across the ensemble. 
With the mode matrix $U=[U_1,\ldots,U_K]$, $\Gamma$ induces a collective-displacement covariance $U\Gamma U^\top$, capturing ensemble-level variation in the learned mode subspace.
During training, we supervise $\Gamma$ against the empirical covariance of MD-projected coefficients.

\subsection{Conditioning Interface}
\label{ci}
Given the learned collective modes $U$ and a coefficient vector $c$, we construct the displacement field $d(c)$ and derive residue-level and pairwise conditioning features encoding local displacement amplitudes and coordinated inter-residue variation.
For residue $i$ and mode $k$, $\|U_{i,k}\||c_k|$ captures mode-specific magnitude, while $\tilde d_i(c)=r_i^{\top}d_i(c)$ gives rotation-invariant local coordinates while preserving direction.
For each residue pair $(i,j)$, we encode the mode-resolved coordination $\langle U_{i,k},U_{j,k}\rangle c_k^2$ and the aggregate displacement relation $\langle d_i(c),d_j(c)\rangle$, which characterize coordinated motion at the individual-mode and combined-displacement levels, respectively.
These features are combined with the structural representation $h$ and injected into the generative network.
To complement feature-level conditioning, we recenter the translational prior using the collective displacement $d(c)$:
\begin{equation}
z^{(0)} = z_{\mathrm{eq}} + d(c) + \sigma_z \epsilon_z, \qquad \epsilon_z \sim \mathcal{N}(0,I),
\end{equation}
where $\sigma_z$ controls the translational noise scale.
Prior recentering is applied only to translations, while orientations are handled by the SE(3) flow.
During training, each MD target $x$ is conditioned on its projected coefficient vector $c^\ast(x)$, used to construct the conditioning features and displacement $d(c^\ast(x))$.
At inference time, we instead sample $c\sim q(c\mid x_{\mathrm{eq}})$ and apply the same conditioning interface to generate a conformation associated with the sampled collective variation.

\subsection{Training Objective}
For each MD target $x$, we sample $\tau\sim\mathcal U(0,1)$ and train the SE(3) flow along a conditional path from the coefficient-conditioned base state to $x$, using linear interpolation for translations and geodesic interpolation for rotations.
The flow objective is:
\begin{equation}
\mathcal{L}_{\mathrm{flow}}=\mathbb{E}\!\left[\frac{1}{\sum_i w_i}\sum_i w_i\left(\|v_{\theta,i}^{z}-u_{i,z}^{\ast}\|_2^2+\|v_{\theta,i}^{r}-u_{i,r}^{\ast}\|_{\mathrm{SO}(3)}^2\right)\right],
\end{equation}
where $v_{\theta,i}^{z}$ is the predicted translational velocity and $v_{\theta,i}^{r}\in T_{r_i^{(\tau)}}\mathrm{SO}(3)$ is the predicted rotational tangent vector.
The target velocities $u_{i,z}^{\ast}$ and $u_{i,r}^{\ast}$ are defined in \autoref{sec:implementation_details}, and the rotational error is measured using the standard $\mathrm{SO}(3)$ metric.
For each protein, we compute the empirical covariance of the MD-projected coefficients over all retained training conformations:
\begin{equation}
\bar c^{\ast}=\frac{1}{T_p}\sum_{t=1}^{T_p}c^{\ast(t)},\qquad
C_c=\frac{1}{T_p}\sum_{t=1}^{T_p}\left(c^{\ast(t)}-\bar c^{\ast}\right)\left(c^{\ast(t)}-\bar c^{\ast}\right)^\top,
\end{equation}
where $c^{\ast(t)}=c^\ast(x_{\mathrm{MD}}^{(t)})$ and $T_p$ is the number of retained training conformations pooled across all MD replicas.
At each $\mathcal{L}_{\mathrm{ens}}$ evaluation, $C_c$ is recomputed from all retained training conformations under the current $U$, with gradients through the projection.
We align $\Gamma$ with $C_c$ via:
\begin{equation}
\mathcal{L}_{\mathrm{ens}} = \frac{\left\|\Gamma-C_c\right\|_F^2} {\left\|C_c\right\|_F^2+\epsilon}.
\end{equation}
The \textbf{objective} is
$\mathcal{L}_{\mathrm{total}}
=\mathcal{L}_{\mathrm{flow}}
+\lambda_{\mathrm{rec}}\mathcal{L}_{\mathrm{rec}}
+\lambda_{\mathrm{coeff}}\mathcal{L}_{\mathrm{coeff}}
+\lambda_{\mathrm{ens}}\mathcal{L}_{\mathrm{ens}}$.
$\lambda$ terms are loss weight.

\section{Experiment}
\subsection{Experimental Setting}
\textbf{Datasets.}
We evaluate all models on two complementary benchmarks with all-atom MD trajectories.
\textbf{ATLAS} \citep{vander2024atlas} contains three independent 100-ns MD replicas for 1,390 diverse natural proteins simulated under standardized conditions.
Following previous work \citep{jing2024alphafold,wolf2025conformational}, we split ATLAS into 1,269 training proteins, 39 validation proteins, and 82 test proteins.
\textbf{de novo proteins} \citep{wolf2025conformational} is a fully held-out benchmark of 100 computationally designed proteins, each with three independent 100-ns MD replicas.
It is used exclusively for out-of-distribution evaluation, testing generalization from natural to designed proteins with little or no informative evolutionary signal.

\textbf{Evaluation Metrics.}
We evaluate RMSF, PwRMSD, PCA, DCCM, and transient-contact Jaccard $J_{\mathrm{tr}}$, measuring residue-wise flexibility, conformational spread, dominant collective motions, correlated residue motions, and recovery of non-equilibrium contacts, respectively.
RMSF and DCCM use Pearson correlation $r$, RMSF and PwRMSD use MAE, and PCA uses the Wasserstein-2 distance $W_2$ between generated and MD distributions projected onto the first two MD principal components.

\noindent\textbf{Baselines.} 
We compare CoMoFlow with representative conformational ensemble generation methods from three categories: pretrained folding-model-based methods, including AlphaFlow, AlphaFlow-T, ESMFlow-T, AlphaFlow-T${\mathrm{dist}}$, AlphaFlow-T${\mathrm{12L,dist}}$, and ConfDiff \citep{wang2024proteinconfdiff}, and P2DFlow \citep{jin2025p2dflow}; the from-scratch structure-conditioned method BBFlow \citep{wolf2025conformational}; and the large-scale pretrained ensemble generator BioEmu \citep{lewis2025scalable}.
Further dataset, evaluation, model, and training details are provided in \autoref{sec:data_eval_details} and \autoref{sec:implementation_details}.

\begin{table*}[t]
\caption{Performance comparison on the ATLAS and de novo proteins datasets.
The reference RMSF medians (Med.) are \textbf{1.48} for ATLAS and \textbf{0.91} for de novo proteins.}
\begin{center}
\setlength{\tabcolsep}{2.3pt}
\renewcommand{\arraystretch}{0.85}
\small
\label{main_results}
\begin{tabular}{@{}llcccccccc@{}}
\toprule
& 
& \multicolumn{3}{c}{RMSF}
& \multicolumn{1}{c}{PwRMSD}
& \multicolumn{1}{c}{DCCM}
& \multirow{2}{*}{PCA $W_2$ $\downarrow$}
& \multirow{2}{*}{$J_{\mathrm{tr}}$ (\%) $\uparrow$}
& \multirow{2}{*}{Time [s] $\downarrow$} \\
\cmidrule(lr){3-5}
\cmidrule(lr){6-6}
\cmidrule(lr){7-7}
Dataset
& Method
& $r$ $\uparrow$
& MAE $\downarrow$
& Med.
& MAE $\downarrow$
& $r$ $\uparrow$
& & & \\
\midrule

\multirow{8}{*}{ATLAS}
& BioEmu
& 0.83 & 1.29 & 2.34 & 2.84 & 0.80 & 1.65 & 36 & 1.9 \\
& AlphaFlow
& 0.86 & 0.59 & 1.51 & 1.35 & 0.86 & 1.47 & 41 & 32.0 \\
& ConfDiff
& 0.88 & 0.62 & 2.00 & 1.45 & 0.86 & 1.41 & 39 & 20.2 \\
& AlphaFlow-T
& \textbf{0.92} & 0.41 & 1.17 & 0.91 & 0.89 & 1.28 & \textbf{47} & 32.6 \\
& ESMFlow-T
& \textbf{0.92} & 0.52 & 0.94 & 1.22 & 0.89 & 1.48 & \textbf{47} & 11.2 \\
& AlphaFlow-T$_{\mathrm{dist}}$
& \textbf{0.92} & 0.68 & 0.90 & 1.41 & 0.88 & 1.43 & 42 & 3.3 \\
& AlphaFlow-T$_{\mathrm{12L,dist}}$
& 0.90 & 0.85 & 0.68 & 1.80 & 0.87 & 1.60 & 24 & 1.2 \\
& P2DFlow
& 0.88 & 0.53 & 1.43 & 0.92 & 0.88 & 1.30 & 33 & 11 \\
& BBFlow
& 0.90 & 0.42 & 1.49 & 0.77 & 0.87 & 1.33 & 29 & \textbf{0.8} \\
\cmidrule(lr){2-10}
& CoMoFlow
& \textbf{0.92} & \textbf{0.38} & \textbf{1.48} & \textbf{0.71}
& \textbf{0.93} & \textbf{1.21} & 37 & \textbf{0.8} \\

\midrule

\multirow{9}{*}{de novo}
& BioEmu
& 0.60 & 4.24 & 7.56 & 8.29 & 0.64 & 1.53 & 23 & 1.9 \\
& AlphaFlow
& 0.47 & 4.76 & 7.09 & 7.40 & 0.58 & 1.64 & 17 & 32.0 \\
& ConfDiff
& 0.62 & 3.82 & 6.35 & 7.26 & 0.65 & 1.72 & 15 & 20.2 \\
& AlphaFlow-T
& \textbf{0.89} & 0.25 & 0.74 & 0.38 & 0.85 & 0.66 & 55 & 32.6 \\
& ESMFlow-T
& \textbf{0.89} & 0.28 & 0.68 & 0.43 & 0.86 & 0.63 & \textbf{56} & 11.2 \\
& AlphaFlow-T$_{\mathrm{dist}}$
& 0.88 & 0.46 & 0.51 & 0.77 & 0.84 & 0.69 & 51 & 3.3 \\
& AlphaFlow-T$_{\mathrm{12L,dist}}$
& 0.87 & 0.58 & 0.41 & 0.97 & 0.83 & 0.75 & 38 & 1.2 \\
& P2DFlow
& 0.80 & 0.42 & 1.05 & 0.56 & 0.79 & 0.83 & 34 & 11 \\
& BBFlow
& 0.84 & 0.26 & 0.87 & 0.32 & 0.83 & 0.67 & 32 & \textbf{0.8} \\
\cmidrule(lr){2-10}
& CoMoFlow
& 0.86 & \textbf{0.23} & \textbf{0.93} & \textbf{0.27}
& \textbf{0.89} & \textbf{0.59} & 43 & \textbf{0.8} \\
\bottomrule
\end{tabular}
\end{center}
\vspace{-8mm}
\end{table*}

\subsection{Basic Performance}

\textbf{ATLAS.}
As the upper part of \autoref{main_results} shows, CoMoFlow outperforms compared methods on most metrics while maintaining high inference efficiency.
Compared with BBFlow, CoMoFlow improves RMSF correlation from 0.90 to 0.92 and MAE from 0.42 to 0.38~\AA, while matching the MD median RMSF of 1.48~\AA.
PwRMSD MAE decreases from 0.77 to 0.71~\AA.
More notably, DCCM correlation improves from 0.87 to 0.93 and PCA $W_2$ decreases from 1.33 to 1.21, indicating better recovery of coordinated and global conformational variation.

\textbf{de novo proteins.}
As shown in the lower part of \autoref{main_results}, CoMoFlow performs better on most metrics, demonstrating generalization to unseen proteins.
Compared with BBFlow, CoMoFlow improves RMSF correlation from 0.84 to 0.86, reduces RMSF MAE from 0.26 to 0.23~\AA, and PwRMSD MAE from 0.32 to 0.27~\AA.
More notably, DCCM correlation improves from 0.83 to 0.89 and PCA $W_2$ decreases from 0.67 to 0.59, indicating better recovery of coordinated and global conformational variation.
Overall, improvements on both benchmarks support the effectiveness of explicit collective modeling across different protein distributions.

\textbf{Ablation Studies.}
As shown in \autoref{tab:ablation}, removing both conditioning pathways reduces DCCM correlation from 0.93 to 0.87 and increases PCA $W_2$ from 1.21 to 1.38, while either pathway alone recovers part of the loss, indicating complementary effects.
Removing $\mathcal{L}_{\mathrm{ens}}$ largely preserves RMSF but degrades DCCM and PCA $W_2$, supporting the role of cross-mode covariance calibration in coordinated variation.
Removing both $\mathcal{L}_{\mathrm{rec}}$ and $\mathcal{L}_{\mathrm{ens}}$ approaches the no-guidance variant, showing that both supervision levels anchor the learned modes to MD variation.
Further analyses of the coefficient distribution and covariance parameterization are provided in \autoref{sec:coefficient_distribution} and \autoref{sec:full_covariance}.

\subsection{Coordinated Variation Recovery}
\begin{figure*}[t]
    \centering
    \includegraphics[width=\textwidth]{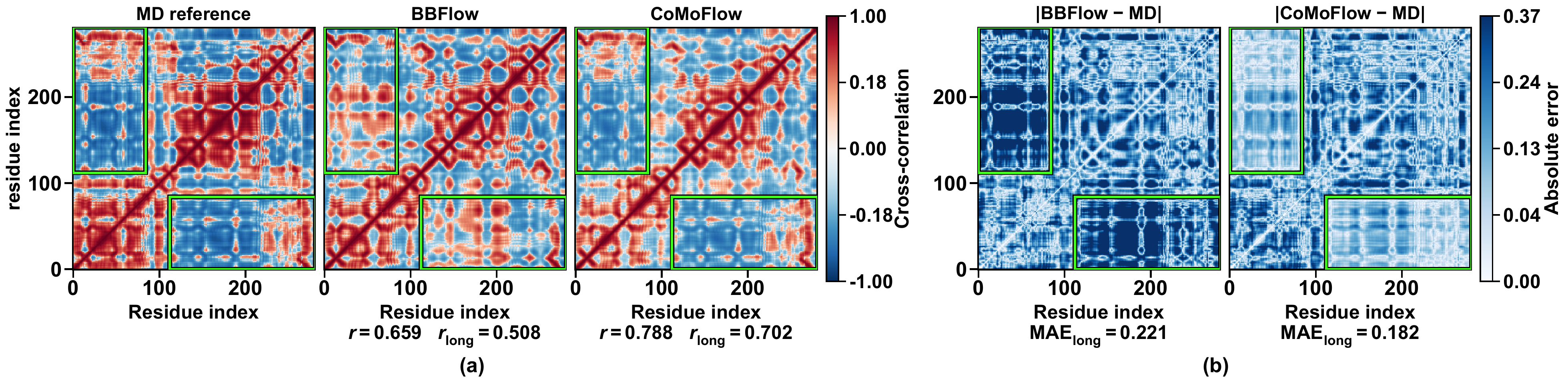}
    \vspace{-8mm}
    \caption{
    Visualization of coordinated conformational variation recovery for protein 7asgA.
    \textbf{(a)} Residue-wise DCCMs of the MD reference and generated ensembles.
    \textbf{(b)} Absolute errors.
    }
    \label{fig:dccm}
    \vspace{-2mm}
\end{figure*}

To further examine the coordinated structural variation underlying the aggregate DCCM results, we compare residue-wise correlation patterns for representative proteins.
As shown in \autoref{fig:dccm}(a), CoMoFlow more closely recovers the correlation patterns observed in the MD ensemble, particularly in the highlighted long-range regions.
For the representative protein, the overall DCCM correlation increases from $0.659$ with BBFlow to $0.788$ with CoMoFlow, while the long-range correlation improves from $0.508$ to $0.702$.
Consistently, \autoref{fig:dccm}(b) shows that the long-range absolute error decreases from $0.221$ to $0.182$.
These results show that the improvement in aggregate DCCM metrics arises from more accurate recovery of coordinated variation across distant residues.

\begin{table*}[t]
\caption{Ablation of collective-conditioning pathways and training objectives on the ATLAS dataset.}
\begin{center}
\setlength{\tabcolsep}{4pt}
\renewcommand{\arraystretch}{0.85}
\small
\label{tab:ablation}
\begin{tabular}{@{}lccccccc@{}}
\toprule
& \multicolumn{3}{c}{RMSF}
& \multicolumn{1}{c}{PwRMSD}
& \multicolumn{1}{c}{DCCM}
& \multirow{2}{*}{PCA $W_2$ $\downarrow$}
& \multirow{2}{*}{$J_{\mathrm{tr}}$ (\%) $\uparrow$} \\
\cmidrule(lr){2-4}
\cmidrule(lr){5-5}
\cmidrule(lr){6-6}
Variant
& $r$ $\uparrow$
& MAE $\downarrow$
& Med. (1.48)
& MAE $\downarrow$
& $r$ $\uparrow$
& & \\
\midrule
w/o collective guidance
& 0.89 & 0.47 & 1.50 & 0.77 & 0.87 & 1.38 & 29 \\
Feature conditioning only
& 0.91 & 0.43 & 1.47 & 0.74 & 0.88 & 1.33 & 31 \\
Prior recentering only
& 0.90 & 0.41 & 1.46 & 0.75 & 0.90 & 1.29 & 32 \\
w/o $\mathcal{L}_{\mathrm{rec}}$
& 0.91 & 0.41 & 1.46 & 0.74 & 0.90 & 1.28 & 33 \\
w/o $\mathcal{L}_{\mathrm{coeff}}$
& 0.90 & 0.42 & 1.44 & 0.76 & 0.89 & 1.31 & 31 \\
w/o $\mathcal{L}_{\mathrm{ens}}$
& 0.91 & 0.39 & 1.48 & 0.72 & 0.91 & 1.25 & 34 \\
w/o $\mathcal{L}_{\mathrm{rec}},\mathcal{L}_{\mathrm{ens}}$
& 0.90 & 0.42 & 1.45 & 0.76 & 0.88 & 1.32 & 30 \\
\midrule
\textbf{CoMoFlow} & \textbf{0.92} & \textbf{0.38} & \textbf{1.48} & \textbf{0.71} & \textbf{0.93} & \textbf{1.21} & \textbf{37} \\
\bottomrule
\end{tabular}
\end{center}
\vspace{-5mm}
\end{table*}

\vspace{-2mm}
\subsection{Controllable Conformation Generation}
\label{sec:ccg}
\textbf{Control Evaluation Protocol.} 
We evaluate whether the learned coefficients provide direct control over generated collective variations.
For each test protein, we sample $c\sim q(c\mid x_{\mathrm{eq}})$, vary one target coefficient $c_k$ around its sampled value, and fix the others.
After alignment to the equilibrium structure, we project the generated displacement onto $U_k$ to obtain the \textbf{realized coefficient}, $\hat{c}_k=\langle z_{\mathrm{gen}}-z_{\mathrm{eq}},U_k\rangle_w$.
We evaluate control fidelity using correlation, regression slope, and normalized MAE (NMAE) between the prescribed and realized coefficient changes.
Correlation measures response consistency, while slope and NMAE measure magnitude fidelity.
We measure off-target variation along remaining modes to assess mode specificity.
See \autoref{sec:control_metrics} for metric definitions.

\textbf{Quantitative Control Fidelity.}
As shown in \autoref{tab:control_fidelity}, CoMoFlow achieves correlations of 0.983 and 0.971 on ATLAS and de novo proteins, respectively, indicating highly consistent control responses.
The corresponding slopes are 0.864 and 0.797, with NMAEs of 0.345 and 0.401, indicating that the realized variations also track the prescribed magnitudes.
Meanwhile, the off-target responses remain at 0.32 and 0.22, indicating that manipulating a target coefficient primarily affects its corresponding mode while inducing limited responses along the others.

\begin{wraptable}{r}{0.28\textwidth}
\centering
\vspace{-7mm}
\renewcommand{\arraystretch}{0.85}
\setlength{\tabcolsep}{1.2pt}
\small
\caption{Controllable conformation generation.}
\label{tab:control_fidelity}
\begin{tabular}{@{}lcc@{}}
\toprule
Metric & ATLAS & de novo \\
\midrule
Corr. $\uparrow$
& 0.983 & 0.971 \\
Slope $\rightarrow 1$
& 0.864 & 0.797 \\
NMAE $\downarrow$
& 0.345 & 0.401 \\
Off-target $\downarrow$
& 0.32 & 0.22 \\
\bottomrule
\end{tabular}
\vspace{-4.3mm}
\end{wraptable}

\begin{figure*}[t]
    \centering
    \includegraphics[width=0.95\textwidth]{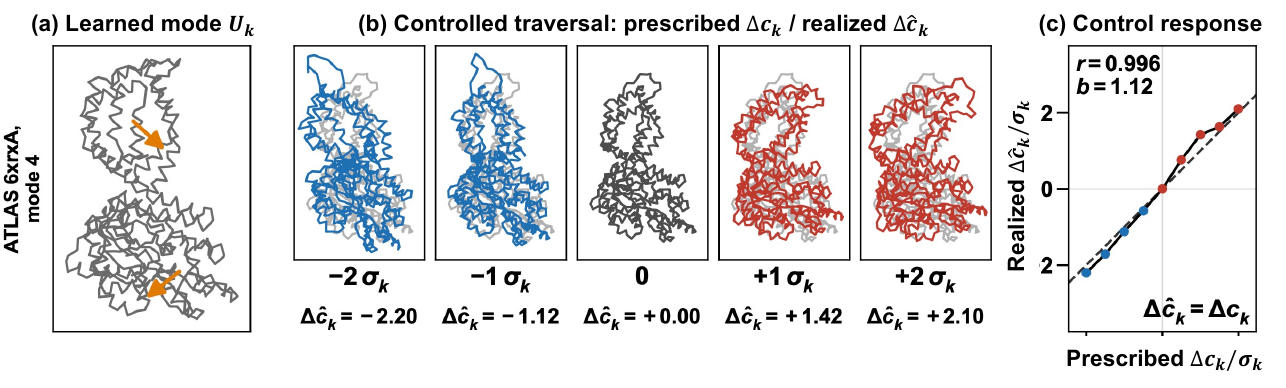}
    \vspace{-6mm}
    \caption{Single-mode controllable generation on a representative protein.}
    \label{fig:control_traversal}
    \vspace{-4mm}
\end{figure*}

\textbf{Single-Mode Traversal.}
To illustrate the structural effects of the learned coefficients, we perform a single-mode traversal.
For a selected mode $U_k$, we vary $\Delta c_k/\sigma_k\in\{-2,-1,0,1,2\}$, fix other coefficients, and generate one conformation per value.
As shown in \autoref{fig:control_traversal}, the generated structures evolve smoothly along $U_k$ as $\Delta c_k$ changes.
Increasing $|\Delta c_k|$ produces larger changes, whereas reversing the sign reverses direction along the same mode.
The realized coefficient changes $\Delta\hat{c}_k$ closely follow the prescribed values, demonstrating direct and interpretable control over collective conformational variations.
The controlled structures retain stable local backbone geometry up to $\pm2\sigma_k$ (\autoref{sec:control_physical}).

\begin{figure*}[t]
    \centering
    \includegraphics[width=1\textwidth]{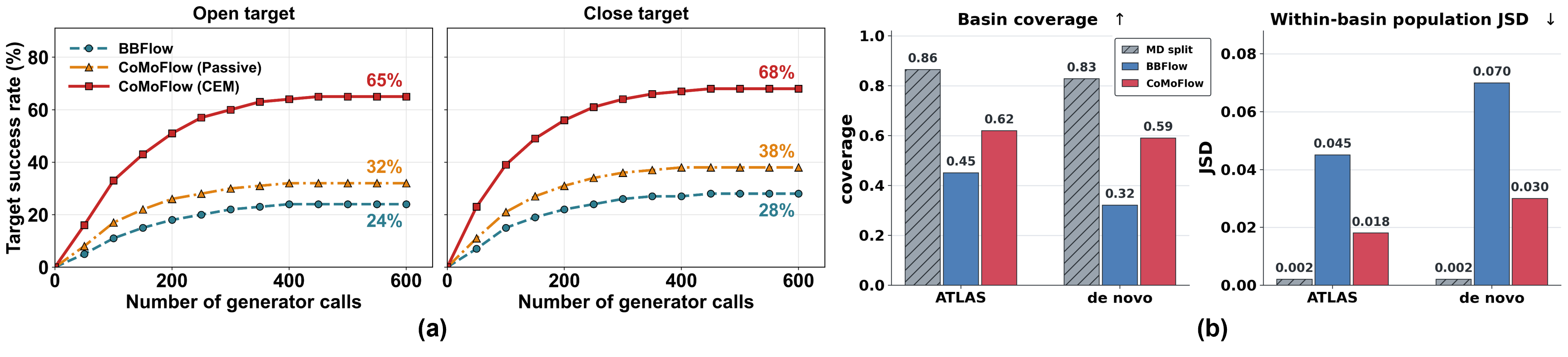}
    \vspace{-8mm}
    \caption{Targeted exploration and conformational-basin recovery. (a) Cumulative opening/closing target success for BBFlow, passive CoMoFlow, and CoMoFlow with CEM. (b) Basin coverage and population JSD for BBFlow and CoMoFlow on ATLAS and de novo proteins.}
    \label{fig:target_exploration}
    \vspace{-4mm}
\end{figure*}

\textbf{Targeted Conformational Exploration.}
We further evaluate target-driven conformational exploration on the ATLAS test set.
For each protein, we select two relatively rigid local regions and define $g(x)$ as their centroid distance, with the MD-derived $P_{95}$ and $P_{5}$ as opening and closing targets.
Under the same budget of 600 generator calls per target, we compare BBFlow using its native passive sampling, CoMoFlow with passive sampling from $q(c\mid x_{\mathrm{eq}})$, and CoMoFlow with CEM search over its explicit low-dimensional collective-mode coefficients.
At each CEM iteration, coefficient vectors are sampled from the current proposal, used to generate candidate conformations, and ranked by $g(x)$, with elite coefficients updating the proposal (10 iterations, 60 candidates per iteration, 20\% elite fraction, $\alpha=0.7$ smoothing).
A target is considered reached if any generated conformation satisfies $g(x)\geq P_{95}$ for opening or $g(x)\leq P_{5}$ for closing.
As shown in \autoref{fig:target_exploration}(a), passive CoMoFlow reaches 32\%/38\% for opening/closing targets versus 24\%/28\% for BBFlow, while CEM further increases the rates to 65\%/68\%, demonstrating both improved target accessibility and the benefit of directly optimizing the explicit collective space.

\subsection{Conformational-Basin Population Recovery}
We further evaluate whether the generated ensembles recover the population structure of distinct conformational regions observed in MD. 
We use conformational basins rather than states, since our analysis identifies structurally separated high-density regions without assuming kinetic metastability. Specifically, aligned MD conformations are clustered with HDBSCAN using C$\alpha$ RMSD, and only basins reproduced in at least two independent MD replicas are retained. 
Generated conformations are assigned to the nearest basin medoid within its 99th-percentile RMSD radius. We evaluate basin coverage and, for proteins with at least two retained basins, the within-basin population JSD between generated and MD ensembles. 
As shown in \autoref{fig:target_exploration}(b), CoMoFlow improves coverage from 0.45 to 0.62 on ATLAS and from 0.32 to 0.59 on de novo proteins, while reducing population JSD from 0.045 to 0.018 and from 0.070 to 0.030, respectively, indicating better recovery of both MD-supported basins and their relative populations.

\section{Conclusion}
This work addresses protein conformational ensemble generation by explicitly modeling collective conformational variations.
We introduce CoMoFlow, which jointly learns protein-specific collective modes and their coefficient distribution, and incorporates sampled variations into SE(3) flow matching for ensemble generation and direct control.
Experiments on ATLAS and de novo proteins demonstrate improved ensemble reconstruction over strong flow-based baselines, while coordinated-variation and control analyses validate the learned collective-mode representation.

\section*{Ethics Statement}
All experiments in this work were conducted entirely in silico using existing molecular dynamics simulation datasets and computationally designed protein benchmarks. 
This study did not involve human participants, animal subjects, clinical data, wet-lab experiments, or the collection of biological specimens. Accordingly, no institutional ethics approval or informed consent was required.

\section*{AI Usage Statement}
Large language models (LLMs) and other text-generative AI systems were not used as part of the scientific methodology or experimental pipeline, including data generation, analysis, evaluation, or interpretation of results. An LLM was used solely to polish the manuscript and correct grammar and spelling. All LLM-assisted edits were reviewed and verified by the authors, who take full responsibility for the final manuscript.

\section*{Reproducibility Statement}
The code associated with this work will be made publicly available upon acceptance. 
Detailed descriptions of the model configurations, training procedures, evaluation protocols, and experimental settings are provided in the paper to support full reproduction of the reported results.

\bibliography{iclr2027_conference}

@article{childers2017insights,
  title={Insights from molecular dynamics simulations for computational protein design},
  author={Childers, Matthew Carter and Daggett, Valerie},
  journal={Molecular systems design \& engineering},
  volume={2},
  number={1},
  pages={9--33},
  year={2017},
  publisher={The Royal Society of Chemistry}
}

@inproceedings{lu2024str2str,
  title={Str2str: A score-based framework for zero-shot protein conformation sampling},
  author={Lu, Jiarui and Zhong, Bozitao and Zhang, Zuobai and Tang, Jian},
  booktitle={International Conference on Learning Representations},
  volume={2024},
  pages={10663--10694},
  year={2024}
}

@article{zheng2024predicting,
  title={Predicting equilibrium distributions for molecular systems with deep learning},
  author={Zheng, Shuxin and He, Jiyan and Liu, Chang and Shi, Yu and Lu, Ziheng and Feng, Weitao and Ju, Fusong and Wang, Jiaxi and Zhu, Jianwei and Min, Yaosen and others},
  journal={Nature Machine Intelligence},
  volume={6},
  number={5},
  pages={558--567},
  year={2024},
  publisher={Nature Publishing Group UK London}
}

@article{janson2025deep,
  title={Deep generative modeling of temperature-dependent structural ensembles of proteins},
  author={Janson, Giacomo and Jussupow, Alexander and Feig, Michael},
  journal={Communications Chemistry},
  volume={8},
  number={1},
  pages={354},
  year={2025},
  publisher={Nature Publishing Group UK London}
}

@article{sugita1999replica,
  title={Replica-exchange molecular dynamics method for protein folding},
  author={Sugita, Yuji and Okamoto, Yuko},
  journal={Chemical physics letters},
  volume={314},
  number={1-2},
  pages={141--151},
  year={1999},
  publisher={Elsevier}
}

@article{karplus2002molecular,
  title={Molecular dynamics simulations of biomolecules},
  author={Karplus, Martin and McCammon, J Andrew},
  journal={Nature structural biology},
  volume={9},
  number={9},
  pages={646--652},
  year={2002},
  publisher={Nature Publishing Group US New York}
}

@inproceedings{lipman2023flow,
title={Flow Matching for Generative Modeling},
author={Yaron Lipman and Ricky T. Q. Chen and Heli Ben-Hamu and Maximilian Nickel and Matthew Le},
booktitle={The Eleventh International Conference on Learning Representations },
year={2023},
url={https://openreview.net/forum?id=PqvMRDCJT9t}
}

@article{amadei1993essential,
  title={Essential dynamics of proteins},
  author={Amadei, Andrea and Linssen, Antonius BM and Berendsen, Herman JC},
  journal={Proteins: Structure, Function, and Bioinformatics},
  volume={17},
  number={4},
  pages={412--425},
  year={1993},
  publisher={Wiley Online Library}
}

@article{garcia1992large,
  title={Large-amplitude nonlinear motions in proteins},
  author={Garc{\'\i}a, Angel E},
  journal={Physical review letters},
  volume={68},
  number={17},
  pages={2696},
  year={1992},
  publisher={APS}
}

@inproceedings{li2024improving,
title={Improving AlphaFlow for Efficient Protein Ensembles Generation},
author={Shaoning Li and Mingyu Li and Yusong Wang and Xinheng He and Zhang Jian and Nanning Zheng and Pheng-Ann Heng},
booktitle={ICML 2024 AI for Science Workshop},
year={2024},
url={https://openreview.net/forum?id=Z6fPAsu91p}
}

@article{sengar2026generative,
  title={Generative modeling of full-atom protein conformations using latent diffusion on graph embeddings},
  author={Sengar, Aditya and Hariri, Ali and Probst, Daniel and Barth, Patrick and Vandergheynst, Pierre},
  journal={Advances in Neural Information Processing Systems},
  volume={38},
  pages={25220--25258},
  year={2026}
}

@article{hollingsworth2018molecular,
  title={Molecular dynamics simulation for all},
  author={Hollingsworth, Scott A and Dror, Ron O},
  journal={Neuron},
  volume={99},
  number={6},
  pages={1129--1143},
  year={2018},
  publisher={Elsevier}
}

@article{atilgan2001anisotropy,
  title={Anisotropy of fluctuation dynamics of proteins with an elastic network model},
  author={Atilgan, Ali Rana and Durell, Stewart R and Jernigan, Robert L and Demirel, Melik C and Keskin, Ozlem and Bahar, Ivet},
  journal={Biophysical journal},
  volume={80},
  number={1},
  pages={505--515},
  year={2001},
  publisher={Elsevier}
}

@article{amaro2010emerging,
  title={Emerging methods for ensemble-based virtual screening},
  author={Amaro, Rommie E and Li, Wilfred W},
  journal={Current topics in medicinal chemistry},
  volume={10},
  number={1},
  pages={3--13},
  year={2010},
  publisher={Bentham Science Publishers direct}
}

@article{sledz2018protein,
  title={Protein structure-based drug design: from docking to molecular dynamics},
  author={{\'S}led{\'z}, Pawe{\l} and Caflisch, Amedeo},
  journal={Current opinion in structural biology},
  volume={48},
  pages={93--102},
  year={2018},
  publisher={Elsevier}
}

@article{lombard2025seamoon,
  title={SeaMoon: From protein language models to continuous structural heterogeneity},
  author={Lombard, Valentin and Timsit, Dan and Grudinin, Sergei and Laine, Elodie},
  journal={Structure},
  volume={33},
  number={9},
  pages={1577--1590},
  year={2025},
  publisher={Elsevier}
}

@article{petrone2006can,
  title={Can conformational change be described by only a few normal modes?},
  author={Petrone, Paula and Pande, Vijay S},
  journal={Biophysical journal},
  volume={90},
  number={5},
  pages={1583--1593},
  year={2006},
  publisher={Elsevier}
}

@article{lombard2026petimot,
  title={PETIMOT: a novel framework for inferring protein motions from sparse data using SE (3)-equivariant graph neural networks},
  author={Lombard, Valentin and Van, Julien Nguyen and Grudinin, Sergei and Laine, Elodie},
  journal={Biological Crystallography},
  volume={82},
  number={8},
  year={2026},
  publisher={International Union of Crystallography}
}

@article{bahar1997efficient,
  title={Efficient characterization of collective motions and interresidue correlations in proteins by low-resolution simulations},
  author={Bahar, I and Erman, B and Haliloglu, T and Jernigan, RL},
  journal={Biochemistry},
  volume={36},
  number={44},
  pages={13512--13523},
  year={1997},
  publisher={ACS Publications}
}

@article{tirion1996large,
  title={Large amplitude elastic motions in proteins from a single-parameter, atomic analysis},
  author={Tirion, Monique M},
  journal={Physical review letters},
  volume={77},
  number={9},
  pages={1905},
  year={1996},
  publisher={APS}
}

@article{vander2024atlas,
  title={ATLAS: protein flexibility description from atomistic molecular dynamics simulations},
  author={Vander Meersche, Yann and Cretin, Gabriel and Gheeraert, Aria and Gelly, Jean-Christophe and Galochkina, Tatiana},
  journal={Nucleic acids research},
  volume={52},
  number={D1},
  pages={D384--D392},
  year={2024},
  publisher={Oxford University Press}
}

@article{lewis2025scalable,
  title={Scalable emulation of protein equilibrium ensembles with generative deep learning},
  author={Lewis, Sarah and Hempel, Tim and Jim{\'e}nez-Luna, Jos{\'e} and Gastegger, Michael and Xie, Yu and Foong, Andrew YK and Satorras, Victor Garc{\'\i}a and Abdin, Osama and Veeling, Bastiaan S and Zaporozhets, Iryna and others},
  journal={Science},
  volume={389},
  number={6761},
  pages={eadv9817},
  year={2025},
  publisher={American Association for the Advancement of Science}
}

@inproceedings{wang2024proteinconfdiff,
  title={Protein Conformation Generation via Force-Guided SE (3) Diffusion Models},
  author={Wang, Yan and Wang, Lihao and Shen, Yuning and Wang, Yiqun and Yuan, Huizhuo and Wu, Yue and Gu, Quanquan},
  booktitle={Forty-first International Conference on Machine Learning},
  year={2024}
}

@article{jin2025p2dflow,
  title={P2DFlow: a protein ensemble generative model with SE (3) flow matching},
  author={Jin, Yaowei and Huang, Qi and Song, Ziyang and Zheng, Mingyue and Teng, Dan and Shi, Qian},
  journal={Journal of Chemical Theory and Computation},
  volume={21},
  number={6},
  pages={3288--3296},
  year={2025},
  publisher={ACS Publications}
}

@article{zhu2026extending,
  title={Extending conformational ensemble prediction to multidomain proteins and protein complex},
  author={Zhu, Junjie and B{\"u}low, S{\"o}ren von and Liu, Hongyi and Lindorff-Larsen, Kresten and Chen, Hai-Feng},
  journal={BioRxiv},
  pages={2026--01},
  year={2026},
  publisher={Cold Spring Harbor Laboratory}
}

@inproceedings{jing2024alphafold,
  title={AlphaFold Meets Flow Matching for Generating Protein Ensembles},
  author={Jing, Bowen and Berger, Bonnie and Jaakkola, Tommi},
  year={2024},
  booktitle={Forty-first International Conference on Machine Learning}
}

@article{wolf2025conformational,
  title={Learning conformational ensembles of proteins based on backbone geometry},
  author={Wolf, Nicolas and Seute, Leif and Viliuga, Vsevolod and Wagner, Simon and St{\"u}hmer, Jan and Gr{\"a}ter, Frauke},
  journal={Advances in Neural Information Processing Systems},
  volume={38},
  pages={141926--141957},
  year={2026}
}

@article{henzler2007dynamic,
  title={Dynamic personalities of proteins},
  author={Henzler-Wildman, Katherine and Kern, Dorothee},
  journal={Nature},
  volume={450},
  number={7172},
  pages={964--972},
  year={2007},
  publisher={Nature Publishing Group UK London}
}

@article{boehr2009role,
  title={The role of dynamic conformational ensembles in biomolecular recognition},
  author={Boehr, David D and Nussinov, Ruth and Wright, Peter E},
  journal={Nature chemical biology},
  volume={5},
  number={11},
  pages={789--796},
  year={2009},
  publisher={Nature Publishing Group US New York}
}

@article{tong2023improving,
  title={Improving and generalizing flow-based generative models with minibatch optimal transport},
  author={Tong, Alexander and Fatras, Kilian and Malkin, Nikolay and Huguet, Guillaume and Zhang, Yanlei and Rector-Brooks, Jarrid and Wolf, Guy and Bengio, Yoshua},
  journal={arXiv preprint arXiv:2302.00482},
  year={2023}
}

@inproceedings{wang2509simplefold,
title={SimpleFold: Folding Proteins is Simpler than You Think},
author={Yuyang Wang and Jiarui Lu and Navdeep Jaitly and Joshua M. Susskind and Miguel {\'A}ngel Bautista},
booktitle={The Fourteenth International Conference on Learning Representations},
year={2026},
url={https://openreview.net/forum?id=0j0MmK7EMA}
}

@article{tama2001conformational,
  title={Conformational change of proteins arising from normal mode calculations},
  author={Tama, Florence and Sanejouand, Y-H},
  journal={Protein engineering},
  volume={14},
  number={1},
  pages={1--6},
  year={2001},
  publisher={Oxford University Press}
}

@article{alexandrov2005normal,
  title={Normal modes for predicting protein motions: a comprehensive database assessment and associated Web tool},
  author={Alexandrov, Vadim and Lehnert, Ursula and Echols, Nathaniel and Milburn, Duncan and Engelman, Donald and Gerstein, Mark},
  journal={Protein science},
  volume={14},
  number={3},
  pages={633--643},
  year={2005},
  publisher={Wiley Online Library}
}

@article{yang2007well,
  title={How well can we understand large-scale protein motions using normal modes of elastic network models?},
  author={Yang, Lei and Song, Guang and Jernigan, Robert L},
  journal={Biophysical journal},
  volume={93},
  number={3},
  pages={920--929},
  year={2007},
  publisher={Elsevier}
}

@article{amadei1996efficient,
  title={An efficient method for sampling the essential subspace of proteins},
  author={Amadei, A and Linssen, ABM and de Groot, Bert L and Van Aalten, DMF and Berendsen, HJC},
  journal={Journal of Biomolecular Structure and Dynamics},
  volume={13},
  number={4},
  pages={615--625},
  year={1996},
  publisher={Taylor \& Francis}
}

@article{kitao1998energy,
  title={Energy landscape of a native protein: jumping-among-minima model},
  author={Kitao, Akio and Hayward, Steven and Go, Nobuhiro},
  journal={Proteins: Structure, Function, and Bioinformatics},
  volume={33},
  number={4},
  pages={496--517},
  year={1998},
  publisher={Wiley Online Library}
}

@article{brooks1983harmonic,
  title={Harmonic dynamics of proteins: normal modes and fluctuations in bovine pancreatic trypsin inhibitor.},
  author={Brooks, Bernard and Karplus, Martin},
  journal={Proceedings of the National Academy of Sciences},
  volume={80},
  number={21},
  pages={6571--6575},
  year={1983}
}

@article{hinsen1998analysis,
  title={Analysis of domain motions by approximate normal mode calculations},
  author={Hinsen, Konrad},
  journal={Proteins: Structure, Function, and Bioinformatics},
  volume={33},
  number={3},
  pages={417--429},
  year={1998},
  publisher={Wiley Online Library}
}

@article{hayward1995harmonicity,
  title={Harmonicity and anharmonicity in protein dynamics: a normal mode analysis and principal component analysis},
  author={Hayward, Steven and Kitao, Akio and G{\=o}, Nobuhiro},
  journal={Proteins: Structure, Function, and Bioinformatics},
  volume={23},
  number={2},
  pages={177--186},
  year={1995},
  publisher={Wiley Online Library}
}

@article{levitt1985protein,
  title={Protein normal-mode dynamics: trypsin inhibitor, crambin, ribonuclease and lysozyme},
  author={Levitt, Michael and Sander, Christian and Stern, Peter S},
  journal={Journal of molecular biology},
  volume={181},
  number={3},
  pages={423--447},
  year={1985},
  publisher={Elsevier}
}

@article{bakan2011prody,
  title={ProDy: protein dynamics inferred from theory and experiments},
  author={Bakan, Ahmet and Meireles, Lidio M and Bahar, Ivet},
  journal={Bioinformatics},
  volume={27},
  number={11},
  pages={1575--1577},
  year={2011},
  publisher={Oxford University Press}
}

@inproceedings{bafna2026learning,
  title={Learning residue level protein dynamics with multiscale Gaussians},
  author={Bafna, Mihir and Jing, Bowen and Berger, Bonnie},
  booktitle={International Conference on Learning Representations},
  volume={2026},
  pages={20481--20500},
  year={2026}
}
\bibliographystyle{iclr2027_conference}

\appendix

\section{Dataset Preprocessing and Evaluation Details}
\label{sec:data_eval_details}

\subsection{Dataset Preprocessing}
We follow the preprocessing protocol of BBFlow \citep{wolf2025conformational} for both ATLAS and de novo proteins.
Each protein is represented using the backbone atoms N, C$\alpha$, and C, from which an SE(3) rigid frame is constructed for each residue, with the C$\alpha$ coordinate defining its translation.
Residue identities extracted from the input structure are additionally encoded as sequence features.
For ATLAS, each protein contains three independent MD replicas, each with 1,001 conformations.
During training, one replica is randomly selected and up to 20 conformations are sampled for each protein at each epoch.
These sampled conformations are used for the per-conformation objectives
$\mathcal{L}_{\mathrm{flow}}$, $\mathcal{L}_{\mathrm{rec}}$, and
$\mathcal{L}_{\mathrm{coeff}}$. In contrast, the empirical coefficient covariance
$C_c$ used by $\mathcal{L}_{\mathrm{ens}}$ is computed over all retained training conformations of the corresponding protein under the current learned mode basis.
No temporal striding or systematic frame subsampling is applied.
The effect of reduced MD supervision is evaluated in \autoref{sec:md_efficiency}.
Each MD conformation is centered and aligned to the corresponding equilibrium structure using C$\alpha$-based Kabsch alignment before computing equilibrium-relative displacements.
Proteins longer than 730 residues are excluded from training.
Following previous work, we split ATLAS into 1,269 training proteins, 39 validation proteins, and 82 test proteins.
The de novo proteins benchmark contains 100 proteins and is used exclusively for out-of-distribution evaluation.
At inference time, CoMoFlow only accesses the equilibrium structure and its residue identities; no MD conformations from the test protein are used to infer the collective modes or coefficient distribution.
Robustness to the input reference structure is evaluated in \autoref{sec:reference_robustness}.

\subsection{Ensemble Generation and Evaluation}
For each test protein, we generate 250 conformations using 20 sampling steps with an explicit Euler integrator.
The same generation setting is used for ATLAS and de novo proteins, with the translational noise scale $\sigma_z$ set to $0.90$.
All reported CoMoFlow variants use the same sampling protocol.
Before evaluation, both generated and MD conformations are aligned to the corresponding equilibrium structure.
All metrics reported in the main tables are computed using C$\alpha$ coordinates.
RMSF is measured relative to the equilibrium structure.
RMSF correlation and MAE are first computed for each protein and then summarized by the median across proteins, while the reported RMSF median is computed over the pooled residue-wise RMSF values across proteins.
Pairwise RMSD is computed directly from the aligned C$\alpha$ coordinates without additional pairwise superposition.
For the MD reference, two independently sampled sets of 250 conformations are used to estimate pairwise RMSD.
For PCA evaluation, the principal components are fitted exclusively on the MD ensemble, and both generated and MD conformations are projected onto the first two MD principal components.
The Wasserstein-2 distance $W_2$ is then computed between the resulting projected distributions.
DCCMs are computed from aligned C$\alpha$ coordinates using fluctuations relative to the mean structure of each ensemble.
Transient contacts are evaluated over residue pairs with sequence separation $|i-j|>2$.
A residue pair is considered in contact when its C$\alpha$ distance is below $8\,\text{\AA}$.
Pairs already in contact in the equilibrium structure under the same criterion are excluded.
Among the remaining pairs, a transient contact is defined as one occurring in more than $10\%$ of ensemble conformations.
The transient-contact score $J_{\mathrm{tr}}$ is the Jaccard similarity between the generated and MD transient-contact sets, computed per protein and then averaged across the test set.
All reported results are averaged over five random seeds.

\section{Implementation Details}
\label{sec:implementation_details}
\subsection{Model Configuration}
The collective-mode encoder is a four-layer rotation-invariant message-passing network with a hidden dimension of 128.
For each mode, two scalar heads predict the signed rotational and translational weights $s_{k,i}^{\mathrm{rot}}$ and $s_{k,i}^{\mathrm{trans}}$, while a pooled scalar head predicts the non-negative dimensionless rotational magnitude $\rho_k$.
Pairwise geometric information is encoded using 16 radial basis functions.
We use $K=8$ protein-specific collective modes (Appendix \ref{sec:sensitivity_modes}) and $C=2$ Gaussian mixture components (Appendix \ref{sec:sensitivity_components}) in all main experiments.
The residue-wise residual is predicted as local-frame components by a single linear head from the 128-dimensional residue representation, followed by a $\tanh$ activation and a residual scale of 0.1, and is transformed to the global frame using the corresponding residue orientation.
The SE(3) flow generator uses a six-block Geometric Frame Attention backbone.
The generator uses node and pair dimensions of 240 and 120, respectively.
Collective conditioning is introduced through projected residue- and pair-level adapters before the first generator block, together with residue-level FiLM modulation in the final block.
The residue-wise collective displacement is transformed to the corresponding local frame as
$\tilde d_i(c)=r_i^\top d_i(c)$ before being projected and injected into the generator.

\subsection{Training Configuration}
For flow matching, we sample $\tau\sim\mathcal U(0,1)$.
The translational path and target velocity are defined as:
\begin{equation}
z_i^{(\tau)}=(1-\tau)z_i^{(0)}+\tau z_i^{(1)},\qquad
u_{i,z}^{\ast}=z_i^{(1)}-z_i^{(0)},
\end{equation}
where $z_i^{(0)}$ is the coefficient-recentered translational prior defined in Section \ref{ci}.
For rotations, we follow the equilibrium-conditioned prior of BBFlow. We sample
$r_{i,\mathrm{uncond}}\sim\operatorname{Unif}(\mathrm{SO}(3))$ and set
$r_i^{(0)}=\gamma_{\mathrm{SO}(3)}(r_{i,\mathrm{uncond}},r_{\mathrm{eq},i},\xi)$,
where $\gamma_{\mathrm{SO}(3)}$ denotes geodesic interpolation and $\xi=0.2$.
The rotational path and target velocity are:
\begin{equation}
r_i^{(\tau)}=\operatorname{Exp}_{r_i^{(0)}}\!\left(\tau\operatorname{Log}_{r_i^{(0)}}(r_i^{(1)})\right),\qquad
u_{i,r}^{\ast}=\frac{\operatorname{Log}_{r_i^{(\tau)}}(r_i^{(1)})}{1-\tau}.
\end{equation}
We use the standard $\mathrm{SO}(3)$ metric $\|A\|_{\mathrm{SO}(3)}^2=\frac{1}{2}\operatorname{Tr}(AA^\top)$
for the rotational flow-matching error.
CoMoFlow is trained using AdamW with an initial learning rate of $1\times10^{-4}$ and a weight decay of $0.01$.
We use dynamic batching with a maximum of 16 conformations per GPU, resulting in an effective batch size of 64--128 across eight GPUs, depending on protein length.
Training is performed for 150 epochs with a constant learning rate and no warm-up.
The loss weights are selected based on the validation sensitivity analysis in Appendix \ref{sec:sensitivity_loss}:
\begin{equation}
\lambda_{\mathrm{rec}} = 0.75,
\qquad
\lambda_{\mathrm{coeff}} = 0.5,
\qquad
\lambda_{\mathrm{ens}} = 0.75.
\end{equation}
Models are trained on 8 NVIDIA H800 GPUs.

\section{Definition of Control Metrics}
\label{sec:control_metrics}
For each test protein and each collective mode, we sample five base coefficient vectors
$c^{(0)}\sim q(c\mid x_{\mathrm{eq}})$.
For each base vector and target mode $k$, we evaluate normalized control offsets
$\Delta c_k/\sigma_k\in\{-2,-1,0,1,2\}$ while keeping all non-target coefficients fixed.
The same flow initialization is used across the five offsets within each traversal.
Control metrics are first computed for each traversal, then averaged over base samples and modes within each protein, and finally averaged across test proteins.
For a target mode $k$, let $c^{(0)}$ denote a base coefficient vector and $c^{(s)}$ denote a controlled coefficient vector obtained by varying only $c_k$ while keeping the remaining coefficients fixed.
We define the prescribed normalized response as:
\begin{equation}
z_k^{(s)}=\frac{c_k^{(s)}-c_k^{(0)}}{\sigma_k},
\end{equation}
where $\sigma_k$ denotes the marginal standard deviation of coefficient $k$ under $q(c\mid x_{\mathrm{eq}})$.

After generation, the generated structure is aligned to the equilibrium structure and projected onto the learned collective modes.
The realized coefficient of mode $k$ is:
\begin{equation}
\hat{c}_k^{(s)}=\left\langle z_{\mathrm{gen}}^{(s)}-z_{\mathrm{eq}},U_k\right\rangle_w.
\end{equation}
We then define the normalized realized response relative to the base conformation as:
\begin{equation}
\hat{z}_k^{(s)}=\frac{\hat{c}_k^{(s)}-\hat{c}_k^{(0)}}{\sigma_k}.
\end{equation}

\noindent\textbf{Correlation.}
We compute the Pearson correlation between the prescribed and realized responses:
\begin{equation}
\mathrm{Corr}_k=\operatorname{Pearson}\left(\{z_k^{(s)}\}_s,\{\hat{z}_k^{(s)}\}_s\right).
\end{equation}
A correlation close to one indicates that changes in the prescribed coefficient consistently induce changes in the expected direction.

\noindent\textbf{Slope.}
We measure response magnitude using the regression slope:
\begin{equation}
\mathrm{Slope}_k=\frac{\sum_s z_k^{(s)}\hat{z}_k^{(s)}}{\sum_s\left(z_k^{(s)}\right)^2}.
\end{equation}
A slope close to one indicates that the realized variation closely matches the prescribed magnitude.

\noindent\textbf{Normalized Mean Absolute Error.}
We define the normalized mean absolute error as:
\begin{equation}
\mathrm{NMAE}_k=\frac{\sum_s\left|\hat{z}_k^{(s)}-z_k^{(s)}\right|}{\sum_s\left|z_k^{(s)}\right|+\epsilon}.
\end{equation}
This normalization makes the error comparable across proteins and modes with different coefficient scales.

\noindent\textbf{Off-target Response.}
For each non-target mode $l\neq k$, we define:
\begin{equation}
\hat{z}_l^{(s)}=\frac{\hat{c}_l^{(s)}-\hat{c}_l^{(0)}}{\sigma_l}.
\end{equation}
The off-target response is then measured as:
\begin{equation}
\mathrm{OffTarget}_k=\frac{1}{|\mathcal{S}'|}\sum_{s\in\mathcal{S}'}\frac{\frac{1}{K-1}\sum_{l\neq k}\left|\hat{z}_l^{(s)}\right|}{\left|z_k^{(s)}\right|+\epsilon},
\end{equation}
where $\mathcal{S}'$ excludes the zero-offset control point.
Lower values indicate that varying one coefficient primarily affects its corresponding collective mode while inducing limited variation along the remaining modes.

\begin{table*}[t]
\caption{Sensitivity to the number of collective conformational modes $K$ on the ATLAS and de novo proteins.
The reference RMSF medians (Med.) are \textbf{1.48} for ATLAS and \textbf{0.91} for de novo proteins.}
\begin{center}
\setlength{\tabcolsep}{7.5pt}
\label{tab:sensitivity_k}
\begin{tabular}{@{}llccccccc@{}}
\toprule
&
& \multicolumn{3}{c}{RMSF}
& \multicolumn{1}{c}{PwRMSD}
& \multicolumn{1}{c}{DCCM}
& \multirow{2}{*}{PCA $W_2$ $\downarrow$}
& \multirow{2}{*}{$J_{\mathrm{tr}}$ (\%) $\uparrow$} \\
\cmidrule(lr){3-5}
\cmidrule(lr){6-6}
\cmidrule(lr){7-7}
Dataset
& $K$
& $r$ $\uparrow$
& MAE $\downarrow$
& Med.
& MAE $\downarrow$
& $r$ $\uparrow$
& & \\
\midrule
\multirow{3}{*}{ATLAS}
& 4
& 0.90 & 0.43 & 1.45 & 0.75 & 0.90 & 1.28 & 34 \\
& 8
& \textbf{0.92} & 0.38 & \textbf{1.48} & \textbf{0.71} & \textbf{0.93} & 1.21 & 37 \\
& 16
& \textbf{0.92} & \textbf{0.36} & 1.49 & 0.72 & 0.92 & \textbf{1.19} & \textbf{38} \\
\midrule
\multirow{3}{*}{de novo}
& 4
& 0.84 & 0.27 & 0.89 & 0.30 & 0.86 & 0.64 & 39 \\
& 8
& 0.86 & 0.23 & \textbf{0.93} & \textbf{0.27} & 0.89 & 0.59 & \textbf{43} \\
& 16
& \textbf{0.87} & \textbf{0.22} & \textbf{0.92} & \textbf{0.27} & \textbf{0.90} & \textbf{0.58} & \textbf{43} \\
\bottomrule
\end{tabular}
\end{center}
\end{table*}

\section{Sensitivity Analysis}

\subsection{Sensitivity to the Number of Collective Modes}
\label{sec:sensitivity_modes}
The number of collective modes $K$ determines the dimensionality of the protein-specific conformational representation.
We evaluate CoMoFlow with different numbers of collective modes while keeping all other settings unchanged.
As shown in \autoref{tab:sensitivity_k}, increasing the number of collective modes from $K=4$ to $K=8$ leads to clear improvements on both ATLAS and de novo proteins, particularly in RMSF accuracy, coordinated-motion recovery,
PCA distribution matching, and transient-contact reconstruction.
Further increasing $K$ to 16 yields only modest and metric-dependent changes: some metrics continue to improve slightly, such as RMSF MAE and PCA $W_2$, while others remain nearly unchanged or fluctuate within a narrow
range.
This indicates that most of the dominant collective conformational variation is already captured with $K=8$, with larger mode spaces providing only limited additional benefit.
We therefore use $K=8$ in all main experiments as a compact representation that achieves performance close to that of higher-dimensional mode spaces.

\begin{table*}[t]
\caption{Sensitivity to the number of Gaussian mixture components $C$ on the ATLAS and de novo proteins. 
The reference RMSF medians (Med.) are \textbf{1.48} for ATLAS and \textbf{0.91} for de novo proteins.}
\begin{center}
\setlength{\tabcolsep}{7.5pt}
\label{tab:sensitivity_c}
\begin{tabular}{@{}llccccccc@{}}
\toprule
&
& \multicolumn{3}{c}{RMSF}
& \multicolumn{1}{c}{PwRMSD}
& \multicolumn{1}{c}{DCCM}
& \multirow{2}{*}{PCA $W_2$ $\downarrow$}
& \multirow{2}{*}{$J_{\mathrm{tr}}$ (\%) $\uparrow$} \\
\cmidrule(lr){3-5}
\cmidrule(lr){6-6}
\cmidrule(lr){7-7}
Dataset
& $C$
& $r$ $\uparrow$
& MAE $\downarrow$
& Med.
& MAE $\downarrow$
& $r$ $\uparrow$
& & \\
\midrule
\multirow{3}{*}{ATLAS}
& 1
& 0.91 & 0.42 & 1.46 & 0.74 & 0.90 & 1.27 & 34 \\
& 2
& \textbf{0.92} & 0.38 & \textbf{1.48} & \textbf{0.71} & \textbf{0.93} & 1.21 & \textbf{37} \\
& 4
& \textbf{0.92} & \textbf{0.37} & 1.50 & \textbf{0.71} & 0.91 & \textbf{1.19} & \textbf{37} \\
\midrule
\multirow{3}{*}{de novo}
& 1
& 0.85 & 0.26 & \textbf{0.90} & 0.29 & 0.87 & 0.62 & 40 \\
& 2
& 0.86 & 0.23 & 0.93 & 0.27 & \textbf{0.89} & 0.59 & 43 \\
& 4
& \textbf{0.87} & \textbf{0.22} & \textbf{0.92} & \textbf{0.26} & 0.87 & \textbf{0.58} & \textbf{45} \\
\bottomrule
\end{tabular}
\end{center}
\end{table*}

\subsection{Sensitivity to the Number of Mixture Components}
\label{sec:sensitivity_components}
We next examine the effect of the mixture capacity $C$ in the protein-specific coefficient distribution $q(c\mid x_{\mathrm{eq}})$.
A single Gaussian provides a unimodal approximation to the coefficient distribution, whereas multiple components allow the model to capture more complex joint patterns of collective variation.
We evaluate different numbers of Gaussian mixture components while keeping all other settings unchanged.
As shown in \autoref{tab:sensitivity_c}, replacing the single-Gaussian coefficient model with a mixture distribution generally improves ensemble reconstruction on both ATLAS and de novo proteins.
Increasing the mixture capacity from $C=1$ to $C=2$ yields consistent improvements across most metrics, including RMSF accuracy, pairwise structural variation, DCCM, PCA $W_2$, and transient-contact recovery.
Increasing $C$ further to 4 provides additional gains in several metrics but the improvements are small and not consistent across all measures. 
These results indicate that a small mixture is sufficient to capture the major multimodal variation in the coefficient space, while increasing the mixture capacity beyond $C=2$ provides only limited and metric-dependent benefits.
We therefore use $C=2$ in all main experiments.

\begin{figure*}[t]
    \centering
    \includegraphics[width=\textwidth]{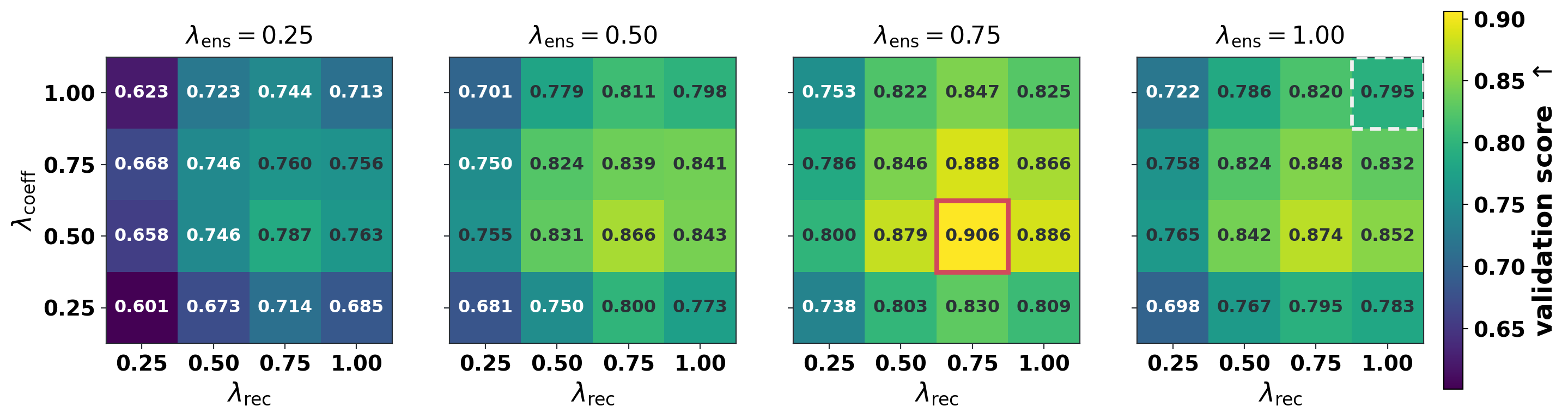}
    \caption{Joint sensitivity analysis of $\lambda_{\mathrm{rec}}$, $\lambda_{\mathrm{coeff}}$, and $\lambda_{\mathrm{ens}}$ on the validation set. Each panel fixes $\lambda_{\mathrm{ens}}$, while varying $\lambda_{\mathrm{rec}}$ and $\lambda_{\mathrm{coeff}}$. The white dashed box highlights the unit-weight setting, where $\lambda_{\mathrm{rec}}=\lambda_{\mathrm{coeff}}=\lambda_{\mathrm{ens}}=1$.}
\end{figure*}

\subsection{Sensitivity to Loss Weights}
\label{sec:sensitivity_loss}
We jointly evaluate the three loss weights $\lambda_{\mathrm{rec}}$, $\lambda_{\mathrm{coeff}}$, and $\lambda_{\mathrm{ens}}$ over $\{0.25,0.50,0.75,1.00\}$, resulting in 64 configurations evaluated on the validation set.
For each configuration, we compute RMSF correlation, RMSF MAE, PwRMSD MAE, DCCM correlation, PCA $W_2$, and transient-contact Jaccard score $J_{\mathrm{tr}}$.
For each metric, we first average over random seeds within each protein and then uniformly across validation proteins, i.e.,
$m_j(\theta)=\frac{1}{P}\sum_{p=1}^{P}\frac{1}{S}\sum_{s=1}^{S}m_j(\theta,p,s)$,
where $P$ and $S$ denote the numbers of validation proteins and random seeds, respectively.
We then convert all metrics to a higher-is-better convention and normalize each metric to $[0,1]$ across the 64 configurations:
\begin{equation}
\tilde{m}_j(\theta)=
\begin{cases}
\dfrac{m_j(\theta)-m_j^{\min}}{m_j^{\max}-m_j^{\min}}, & j\in\mathcal{M}_{\uparrow},\\[6pt]
\dfrac{m_j^{\max}-m_j(\theta)}{m_j^{\max}-m_j^{\min}}, & j\in\mathcal{M}_{\downarrow},
\end{cases}
\end{equation}
where $\theta=(\lambda_{\mathrm{rec}},\lambda_{\mathrm{coeff}},\lambda_{\mathrm{ens}})$,
$\mathcal{M}_{\uparrow}=\{\text{RMSF }r,\text{DCCM }r,J_{\mathrm{tr}}\}$,
and $\mathcal{M}_{\downarrow}=\{\text{RMSF MAE},\text{PwRMSD MAE},\text{PCA }W_2\}$.
Here, $m_j^{\min}$ and $m_j^{\max}$ are the minimum and maximum values of metric $j$ across the 64 configurations.
The composite validation score is the equally weighted mean:
\begin{equation}
S(\theta)=\frac{1}{6}\sum_{j=1}^{6}\tilde{m}_j(\theta).
\end{equation}
As shown in Figure~6, performance remains relatively stable over a broad range of intermediate weights, while weaker or jointly larger weights generally degrade performance.
The best validation score of $0.906$ is obtained at $(\lambda_{\mathrm{rec}},\lambda_{\mathrm{coeff}},\lambda_{\mathrm{ens}})=(0.75,0.50,0.75)$.

\section{Comparison with Classical Collective Modes}
\label{sec:classical_modes}
We compare the learned collective modes in CoMoFlow with classical low-frequency modes obtained from an anisotropic network model (ANM). 
This comparison examines whether classical structure-derived collective modes are sufficient for conformational ensemble generation, or whether learning to predict protein-specific collective modes under MD supervision provides additional benefit.
For each protein, we construct a C$\alpha$-based ANM from the equilibrium structure using a $15$~\AA\ interaction cutoff and a uniform unit spring constant for all residue pairs within the cutoff.
The ANM modes are computed using ProDy.
After removing the six rigid-body modes, we retain the lowest-frequency $K=8$ non-trivial modes, matching the dimensionality used by CoMoFlow.
The resulting modes are represented over the same residues and normalized under the same masked inner product as the learned modes, yielding the fixed classical basis $U_{\mathrm{ANM}}$.
We first evaluate how well each mode subspace captures structural variation in the MD ensemble. 
For each aligned MD conformation, its equilibrium-relative displacement $\Delta z$ is projected onto the corresponding mode basis following Eq.~(7). 
We report the fraction of equilibrium-relative MD displacement energy captured by the $K$-dimensional subspace,
\begin{equation}
\mathrm{Coverage}=1-\frac{\sum_t\left\|\Delta z^{(t)}-\sum_{k=1}^{K}\langle\Delta z^{(t)},U_k\rangle_w U_k\right\|_w^2}{\sum_t\left\|\Delta z^{(t)}\right\|_w^2}.
\end{equation}

To examine whether the choice of mode basis also affects ensemble generation, we construct an \textbf{ANM-CoMoFlow} variant by replacing the learned modes $U$ with the fixed ANM basis $U_{\mathrm{ANM}}$, while keeping the coefficient-distribution model, collective-conditioning interface, SE(3) flow generator, training configuration, and sampling procedure unchanged.
MD conformations are projected onto $U_{\mathrm{ANM}}$ to obtain the target coefficients, and $q(c\mid x_{\mathrm{eq}})$ is learned in the resulting ANM coordinate space.
Since $U_{\mathrm{ANM}}$ is fixed, $\mathcal{L}_{\mathrm{rec}}$ is omitted.
We retain both $\mathcal{L}_{\mathrm{coeff}}$ and $\mathcal{L}_{\mathrm{ens}}$.
For $\mathcal{L}_{\mathrm{ens}}$, the target covariance is computed from MD conformations projected onto the fixed ANM basis, and the predicted coefficient covariance $\Gamma$ is directly matched to this empirical coefficient covariance.

\begin{table*}[t]
\caption{Comparison with classical ANM modes using $K=8$ collective modes.}
\begin{center}
\label{tab:classical_modes}
\begin{tabular*}{0.85\textwidth}{@{\extracolsep{\fill}}llcccc@{}}
\toprule
Dataset & Mode basis & Coverage $\uparrow$ & DCCM $r$ $\uparrow$ & PCA $W_2$ $\downarrow$ & $J_{\mathrm{tr}}$ (\%) $\uparrow$ \\
\midrule
ATLAS
& ANM
& 0.63 & 0.89 & 1.31 & 33 \\
& \textbf{Learned}
& \textbf{0.79} & \textbf{0.93} & \textbf{1.21} & \textbf{37} \\
\midrule
de novo
& ANM
& 0.60 & 0.85 & 0.66 & 39 \\
& \textbf{Learned}
& \textbf{0.76} & \textbf{0.89} & \textbf{0.59} & \textbf{43} \\
\bottomrule
\end{tabular*}
\end{center}
\end{table*}

As shown in \autoref{tab:classical_modes}, classical ANM modes capture a substantial fraction of the MD conformational variation, confirming that low-frequency structure-derived motions provide a meaningful representation of protein flexibility. 
However, the learned CoMoFlow modes consistently capture a larger fraction of the MD displacement energy, with coverage increasing from $0.63$ to $0.79$ on ATLAS and from $0.60$ to $0.76$ on de novo proteins.
This advantage also translates into ensemble generation. 
Compared with ANM-CoMoFlow, the learned modes improve DCCM correlation from $0.89$ to $0.93$ on ATLAS and from $0.85$ to $0.89$ on de novo proteins, while reducing PCA $W_2$ from $1.31$ to $1.21$ and from $0.66$ to $0.59$, respectively. 
Transient-contact recovery follows the same trend. 
These results indicate that classical harmonic modes provide a useful structural prior, whereas the learned protein-specific modes better capture the collective variations expressed across the MD ensemble and consequently provide more effective guidance for conformational ensemble generation.

\section{MD Training-Frame Efficiency}
\label{sec:md_efficiency}
CoMoFlow uses MD conformations during training to supervise the collective-mode representation and coefficient distribution.
We therefore examine how its performance changes when only a subset of the available MD frames is retained for training.
For each ATLAS training protein, we randomly retain $10\%$, $25\%$, $50\%$, or $100\%$ of the available conformations, with the subset fixed throughout training.
The same fraction is applied independently to each of the three MD replicas.
This experiment evaluates training-data efficiency with respect to the number of retained MD frames rather than reduced MD simulation time.
At each epoch, per-conformation training targets are sampled only from the retained subset. The empirical coefficient covariance $C_c$ used by $\mathcal{L}_{\mathrm{ens}}$ is computed over the entire retained subset under the current learned mode basis.
The model architecture, optimization schedule, number of training epochs, and per-epoch sampling procedure remain unchanged.
All models are evaluated against the complete MD ensembles of the ATLAS test set and the fully held-out de novo benchmark.
We evaluate both the quality of the learned collective representation and the resulting ensemble generation.
Mode coverage is computed as defined in \autoref{sec:classical_modes}, while RMSF, PwRMSD, DCCM, PCA $W_2$, and transient-contact accuracy evaluate complementary aspects of conformational ensemble reconstruction.

\begin{table*}[t]
\caption{Effect of MD supervision fraction on conformational ensemble generation. 
The reference RMSF medians (Med.) are \textbf{1.48} for ATLAS and \textbf{0.91} for de novo proteins.}
\begin{center}
\setlength{\tabcolsep}{0.6pt}
\label{tab:md_efficiency}
\begin{tabular}{@{}llcccccccc@{}}
\toprule
&
& \multirow{2}{*}{Coverage $\uparrow$}
& \multicolumn{3}{c}{RMSF}
& \multicolumn{1}{c}{PwRMSD}
& \multicolumn{1}{c}{DCCM}
& \multirow{2}{*}{PCA $W_2$ $\downarrow$}
& \multirow{2}{*}{$J_{\mathrm{tr}}$ (\%) $\uparrow$} \\
\cmidrule(lr){4-6}
\cmidrule(lr){7-7}
\cmidrule(lr){8-8}
Dataset
& MD fraction
&
& $r$ $\uparrow$
& MAE $\downarrow$
& Med.
& MAE $\downarrow$
& $r$ $\uparrow$
& & \\
\midrule

\multirow{4}{*}{ATLAS}
& $10\%$
& 0.69 & 0.88 & 0.47 & 1.42 & 0.79 & 0.88 & 1.36 & 32 \\
& $25\%$
& 0.75 & 0.91 & 0.42 & 1.46 & 0.75 & 0.91 & 1.26 & 36 \\
& $50\%$
& 0.77 & 0.90 & 0.40 & 1.49 & 0.73 & 0.92 & 1.24 & 36 \\
& $100\%$
& \textbf{0.79} & \textbf{0.92} & \textbf{0.38} & \textbf{1.48}
& \textbf{0.71} & \textbf{0.93} & \textbf{1.21} & \textbf{37} \\

\midrule

\multirow{4}{*}{de novo}
& $10\%$
& 0.67 & 0.82 & 0.30 & 0.88 & 0.34 & 0.84 & 0.70 & 38 \\
& $25\%$
& 0.72 & 0.85 & 0.27 & \textbf{0.89} & 0.30 & 0.87 & 0.62 & 41 \\
& $50\%$
& 0.75 & \textbf{0.86} & 0.25 & 0.94 & 0.29 & 0.88 & 0.61 & \textbf{43} \\
& $100\%$
& \textbf{0.76} & \textbf{0.86} & \textbf{0.23} & \textbf{0.93}
& \textbf{0.27} & \textbf{0.89} & \textbf{0.59} & \textbf{43} \\
\bottomrule
\end{tabular}
\end{center}
\end{table*}

As shown in \autoref{tab:md_efficiency}, CoMoFlow remains effective when trained with substantially reduced MD supervision. 
Using only $25\%$ of the available MD conformations already recovers most of the full-supervision performance, while increasing the supervision fraction to $50\%$ further narrows the gap across most metrics. 
On ATLAS, for example, DCCM correlation increases from $0.91$ with $25\%$ supervision to $0.93$ with the full MD set, while PCA $W_2$ decreases from $1.26$ to $1.21$.
A similar overall trend is observed on the fully held-out de novo proteins, indicating that the reduced-supervision models retain much of their generalization performance despite having access to fewer MD conformations during training. 
Although individual metrics exhibit small fluctuations, performance generally improves as more MD supervision is provided and begins to saturate before the full ensemble is used. 
These results suggest that CoMoFlow can learn effective collective conformational representations without requiring the full set of available MD conformations for supervision.

\section{Reference-Structure Robustness}
\label{sec:reference_robustness}
CoMoFlow predicts protein-specific collective modes and their coefficient distribution from a single reference structure.
We therefore examine its sensitivity to the particular reference conformation provided at inference time.
For each test protein, we compute the C$\alpha$ RMSD of all MD conformations from the original equilibrium structure and select alternative reference conformations near the 25th, 50th, and 75th percentiles of this RMSD distribution, denoted as Q25, Q50, and Q75, respectively.
Across the test proteins, the median RMSDs of the alternative references from the original equilibrium structure are $0.74/1.02/1.39$~\AA\ for Q25/Q50/Q75 on ATLAS and $0.56/0.78/1.03$~\AA\ on de novo proteins.

The trained CoMoFlow model is used without retraining or adaptation, with only the input reference structure replaced.
For each alternative reference, the collective modes, coefficient distribution, and coefficient-guided generation are recomputed from that input structure.
For evaluation, generated and MD conformations are aligned to the original equilibrium structure, and RMSF is computed relative to the same original equilibrium structure, ensuring a common evaluation frame across all reference choices.

\begin{table*}[t]
\caption{Effect of alternative reference structures on conformational ensemble generation.
The reference RMSF medians (Med.) are \textbf{1.48} for ATLAS and \textbf{0.91} for de novo proteins.}
\begin{center}
\setlength{\tabcolsep}{4pt}
\label{tab:reference_robustness}
\begin{tabular}{@{}llccccccc@{}}
\toprule
&
& \multicolumn{3}{c}{RMSF}
& \multicolumn{1}{c}{PwRMSD}
& \multicolumn{1}{c}{DCCM}
& \multirow{2}{*}{PCA $W_2$ $\downarrow$}
& \multirow{2}{*}{$J_{\mathrm{tr}}$ (\%) $\uparrow$} \\
\cmidrule(lr){3-5}
\cmidrule(lr){6-6}
\cmidrule(lr){7-7}
Dataset
& Reference
& $r$ $\uparrow$
& MAE $\downarrow$
& Med.
& MAE $\downarrow$
& $r$ $\uparrow$
& & \\
\midrule

\multirow{4}{*}{ATLAS}
& Original
& \textbf{0.92} & \textbf{0.38} & \textbf{1.48} & \textbf{0.71}
& \textbf{0.93} & \textbf{1.21} & \textbf{37} \\
& $Q_{25}$
& 0.90 & 0.42 & 1.43 & 0.76 & 0.90 & 1.29 & 34 \\
& $Q_{50}$
& 0.87 & 0.49 & 1.59 & 0.86 & 0.88 & 1.43 & 35 \\
& $Q_{75}$
& 0.82 & 0.57 & 1.64 & 1.01 & 0.82 & 1.58 & 28 \\

\midrule

\multirow{4}{*}{de novo}
& Original
& \textbf{0.86} & \textbf{0.23} & \textbf{0.93} & \textbf{0.27}
& \textbf{0.89} & \textbf{0.59} & \textbf{43} \\
& $Q_{25}$
& 0.84 & 0.27 & 0.87 & 0.30 & 0.86 & 0.64 & 40 \\
& $Q_{50}$
& 0.81 & 0.32 & 0.99 & 0.37 & 0.87 & 0.74 & 38 \\
& $Q_{75}$
& 0.75 & 0.40 & 1.08 & 0.49 & 0.79 & 0.91 & 32 \\
\bottomrule
\end{tabular}
\end{center}
\end{table*}

As shown in \autoref{tab:reference_robustness}, CoMoFlow is robust to moderate variation in the input reference conformation. 
Using the $Q_{25}$ reference results in only a modest degradation relative to the original equilibrium structure on both ATLAS and de novo proteins. 
As the reference moves further from the equilibrium structure, the overall reconstruction quality deteriorates, with a more pronounced performance drop at $Q_{75}$. 
For example, on ATLAS, DCCM correlation decreases from $0.93$ with the original reference to $0.90$ at $Q_{25}$ and $0.82$ at $Q_{75}$, while PCA $W_2$ increases from $1.21$ to $1.29$ and $1.58$, respectively.
Similar behavior is observed across local flexibility, overall conformational spread, coordinated motion, and ensemble-distribution metrics. 
These results indicate that CoMoFlow does not require a highly specific equilibrium conformation and tolerates moderate structural variation in the reference, while references that deviate substantially from equilibrium lead to a clear loss in ensemble fidelity.

\begin{figure*}[t]
    \centering
    \includegraphics[width=0.95\textwidth]{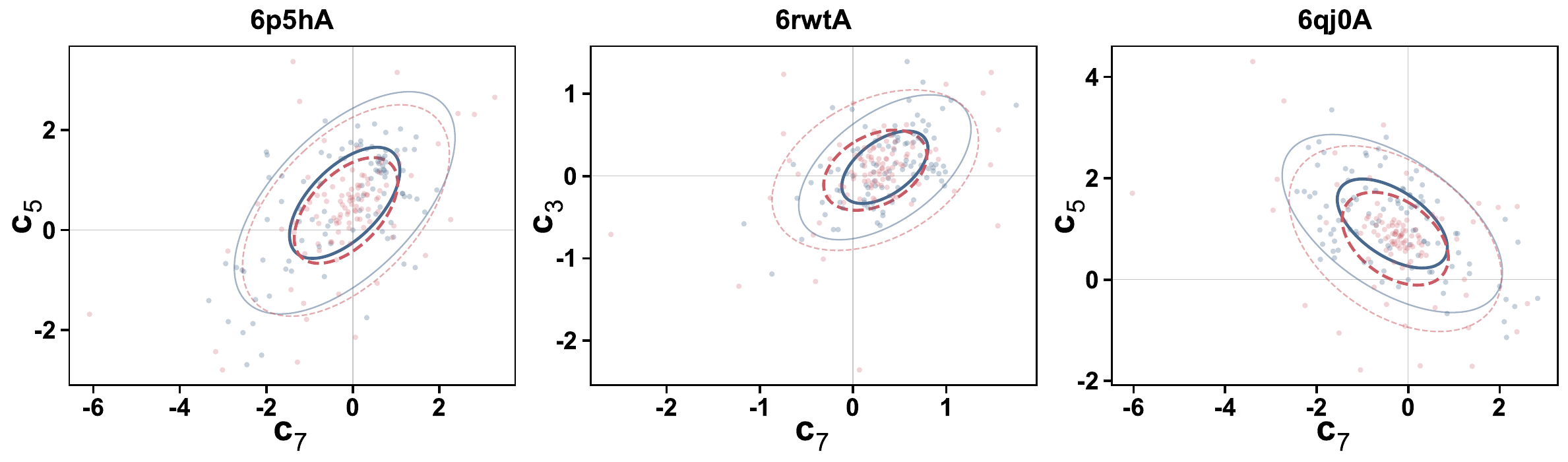}
    \caption{Learned coefficient distributions.
    Joint distributions of MD-projected coefficients (blue) and samples from $q(c\mid x_{\mathrm{eq}})$ (red) along the two highest-variance coefficient dimensions. Ellipses indicate the corresponding covariance structure.}
    \label{fig:target_exploration2}
\end{figure*}

\section{Learned Coefficient Distribution}
\label{sec:coefficient_distribution}
We directly examine whether the predicted distribution $q(c\mid x_{\mathrm{eq}})$ captures the joint statistics of collective-mode coefficients observed in held-out MD conformations.
For each representative test protein, we project its MD conformations onto $U$ to obtain $\{c^{*}(x)\}$ and independently sample coefficients from $q(c\mid x_{\mathrm{eq}})$ using only the equilibrium structure as input. 
For visualization, we select the two coefficient dimensions with the largest variance in the MD-projected ensemble, so that each panel focuses on the dominant coefficient variation of the corresponding protein.
As shown in \autoref{fig:target_exploration2}, the predicted coefficient distributions show similar location, spread, and orientation to the MD-projected coefficient clouds across proteins with distinct patterns of collective variation. 
In particular, the predicted distributions recover both positive and negative cross-mode correlations, indicating that $q(c\mid x_{\mathrm{eq}})$ captures dependencies between collective modes.

\section{Full-Covariance Coefficient Modeling}
\label{sec:full_covariance}
Although the learned collective modes are orthonormal under the masked geometric inner product, their coefficients need not be statistically independent across the conformational ensemble. 
We therefore examine whether explicitly modeling cross-mode dependencies within each Gaussian mixture component is beneficial. 
The default CoMoFlow parameterizes each component with a full covariance matrix $\Sigma_j$, whereas the ablated variant restricts $\Sigma_j$ to be diagonal. 
The two variants are otherwise identical in model architecture, number of mixture components, training objectives, and optimization settings.

We evaluate this design at both the coefficient and ensemble levels.
At the coefficient level, we report the negative log-likelihood (NLL) of the projected MD coefficients under the predicted distribution and the discrepancy between their cross-mode correlation structure and that reproduced by samples from $q(c\mid x_{\mathrm{eq}})$. 
Specifically, for each protein we compute the empirical coefficient correlation matrix $R_{\mathrm{MD}}$ from the projected MD coefficients and $R_q$ from coefficients sampled from the predicted distribution, and measure the mean absolute error over their off-diagonal entries,
\begin{equation}
\mathrm{CorrErr}=\frac{2}{K(K-1)}\sum_{k<l}\left|R_{\mathrm{MD},kl}-R_{q,kl}\right|.
\end{equation}
At the ensemble level, we evaluate coordinated structural variation using DCCM correlation and PCA $W_2$, and basin-level population recovery using basin coverage and population JSD.

\begin{table*}[t]
\caption{Ablation of covariance parameterization in the coefficient distribution.}
\begin{center}
\setlength{\tabcolsep}{2.5pt}
\label{tab:covariance_ablation}
\begin{tabular}{@{}llcccccc@{}}
\toprule
Dataset
& Covariance
& Coeff. NLL $\downarrow$
& CorrErr $\downarrow$
& DCCM $r$ $\uparrow$
& PCA $W_2$ $\downarrow$
& Basin cov. $\uparrow$
& Pop. JSD $\downarrow$ \\
\midrule
\multirow{2}{*}{ATLAS}
& Diagonal
& 9.84 & 0.118 & 0.90 & 1.29 & 0.55 & 0.027 \\
& Full
& \textbf{9.21} & \textbf{0.074} & \textbf{0.93} & \textbf{1.21}
& \textbf{0.62} & \textbf{0.018} \\
\midrule
\multirow{2}{*}{de novo}
& Diagonal
& 9.47 & 0.126 & 0.86 & 0.65 & 0.51 & 0.043 \\
& Full
& \textbf{8.96} & \textbf{0.081} & \textbf{0.89} & \textbf{0.59}
& \textbf{0.59} & \textbf{0.030} \\
\bottomrule
\end{tabular}
\end{center}
\end{table*}

\begin{table*}[t]
\caption{Physical plausibility along controlled single-mode traversals.}
\begin{center}
\setlength{\tabcolsep}{5pt}
\label{tab:control_physical}
\begin{tabular}{@{}llcc@{}}
\toprule
Dataset
& $|\Delta c_k|/\sigma_k$
& C$\alpha$ clash (\%) $\downarrow$
& C$\alpha$--C$\alpha$ bond MAE (\AA) $\downarrow$ \\
\midrule
\multirow{3}{*}{ATLAS}
& 0 & 0.37 & 0.052 \\
& 1 & 0.39 & 0.055 \\
& 2 & 0.42 & 0.061 \\
\midrule
\multirow{3}{*}{de novo}
& 0 & 0.43 & 0.057 \\
& 1 & 0.42 & 0.060 \\
& 2 & 0.47 & 0.066 \\
\bottomrule
\end{tabular}
\end{center}
\end{table*}
As shown in \autoref{tab:covariance_ablation}, the full-covariance model more accurately captures the distribution of collective-mode coefficients.
Compared with the diagonal variant, it achieves lower coefficient NLL and substantially lower CorrErr on both ATLAS and de novo proteins, indicating more faithful recovery of cross-mode statistical dependencies.
This result confirms that geometric orthogonality of the learned modes does not imply statistical independence of their coefficients across the MD ensemble.
The improvement in coefficient modeling also translates into more accurate ensemble reconstruction.
On ATLAS, full covariance improves DCCM correlation from $0.90$ to $0.93$ and reduces PCA $W_2$ from $1.29$ to $1.21$.
It also increases basin coverage from $0.55$ to $0.62$ while reducing population JSD from $0.027$ to $0.018$.
A similar pattern is observed on de novo proteins, where basin coverage increases from $0.51$ to $0.59$ and population JSD decreases from $0.043$ to $0.030$.
These results show that explicitly modeling cross-mode dependencies improves not only the latent coefficient distribution, but also the recovery of coordinated conformational variation and the population structure of the generated ensemble.

\section{Physical Plausibility of Controlled Traversals}
\label{sec:control_physical}
We further examine whether coefficient-controlled traversal introduces local geometric distortions.
For each test protein, we perform the same single-mode traversal as in Section \ref{sec:ccg} and evaluate generated conformations at $|\Delta c_k|/\sigma_k \in \{0,1,2\}$.
We report two lightweight backbone-geometry metrics.
The C$\alpha$ clash rate is defined as the fraction of nonlocal residue pairs with sequence separation greater than two whose C$\alpha$ distance is below $3.0$~\AA.
The virtual-bond deviation is the mean absolute deviation of consecutive C$\alpha$--C$\alpha$ distances from $3.8$~\AA.
As shown in \autoref{tab:control_physical}, both metrics remain stable as the control magnitude increases.
Even at $|\Delta c_k|=2\sigma_k$, the increase in C$\alpha$ clashes is small and consecutive-residue geometry remains close to the expected backbone spacing.
These results indicate that the controllable structural changes are not achieved through substantial local geometric distortion.

\end{document}